\documentclass[]{aero}

\usepackage{tabularx}
\usepackage[flushleft]{threeparttable}

\usepackage{algorithm}
\usepackage{algpseudocode}

\usepackage{hf-tikz} 
\usetikzlibrary{tikzmark,positioning,shapes,arrows.meta}

\ADsetup{
title    = {Energy efficiency analysis of Spiking Neural Networks for space applications},
author   = {Paolo Lunghi$^{1}$\cor{}, Stefano Silvestrini$^{1}$, Dominik Dold$^{2}$, Gabriele Meoni$^{2, 3}$, Alexander Hadjiivanov$^{2}$ and Dario Izzo$^{2}$
           },
address  = {
  1. Politecnico di Milano, Department of Aerospace Science and Technology, Via La Masa 34, 20156, Milano, IT \sep
  2. European Space Agency, Advanced Concepts Team, Keplerlaan 1, 2201 AZ Noordwijk, NL \sep
  3. European Space Agency, $\Phi$-lab, Via Galileo Galilei 1, 00044 Frascati, Italy
},
email    = {paolo.lunghi@polimi.it},
abstract = {
  While the exponential growth of the space sector and new operative concepts ask for higher spacecraft autonomy, the development of AI-assisted space systems was so far hindered by the low availability of power and energy typical of space applications.
In this context, Spiking Neural Networks (SNN) are highly attractive due to their theoretically superior energy efficiency due to their inherently sparse activity induced by neurons communicating by means of binary spikes.
Nevertheless, the ability of SNN to reach such efficiency on real world tasks is still to be demonstrated in practice.
To evaluate the feasibility of utilizing SNN onboard spacecraft, this work presents a numerical analysis and comparison of different SNN techniques applied to scene classification for the EuroSAT dataset.
Such tasks are of primary importance for space applications and constitute a valuable test case given the abundance of competitive methods available to establish a benchmark.
Particular emphasis is placed on models based on temporal coding, where crucial information is encoded in the timing of neuron spikes.
These models promise even greater efficiency of resulting networks, as they maximize the sparsity properties inherent in SNN.
A reliable metric capable of comparing different architectures in a hardware-agnostic way is developed to establish a clear theoretical dependence between architecture parameters and the energy consumption that can be expected onboard the spacecraft.
The potential of this novel method and his flexibility to describe specific hardware platforms is demonstrated by its application to predicting the energy consumption of a BrainChip Akida AKD1000 neuromorphic processor.

  },
keywords = {
  Spiking Neural Networks \sep Spacecraft Autonomy \sep Energy Consumption \sep Interdisciplinary Research 
},
}

\begin{document}

\maketitle  

\section{Introduction}
\label{sec:intro}
New operative concepts such as in-orbit servicing, active debris removal, megaconstellations, interplanetary cubesats, space traffic control, and reusability of space assets are leading a change of paradigm in the space sector, with unprecedented growth in the number of spacecraft in flight and increasing complexity in mission design.
However, several bottlenecks persist that hold back this paradigm shift, particularly the limited access of satellites to the ground segment, mostly constrained by the use of large communication structures and limited transit windows where communication is possible.
Nevertheless, the current state of the art is still to guide space missions from ground with extensive human intervention \cite{harris_spacecraft_2019-1}. 
Although this is a reliable process, it represents a high cost for a mission \cite{saing_nasa_2020-1}, and the development of ground infrastructure is struggling to cope with the growing needs \cite{published_spacex_2023,diaz_data-driven_2023}.
All these factors clearly demand a higher level of \textit{system autonomy} and capability to adapt to different working conditions \cite{moghaddam_guidance_2021,luu_review_2021,querejazu_small_2017}.
Thus, it is not surprising that in recent years, interest in artificial intelligence (AI) applications onboard satellites has grown substantially (see \cite{izzo2019survey, izzo2022selected} for an overview).
Potential applications in space systems include, among others, early detection of potential failures or catastrophic events \cite{giuffrida_cloudscout_2020}, feature detection and tracking in Guidance, Navigation, and Control systems \cite{bechini_robust_2024,bechini_dataset_2023,silvestrini_implicit_2022,silvestrini_optical_2022}, pre-processing of collected data by Earth Observation satellites to mitigate bandwidth requirements by filtering out invalid or corrupted data, advanced navigation and signal processing tasks \cite{meoni2024ops,revach_kalmannet_2022,niitsoo_deep_2019}.
On the other hand, space systems are traditionally limited in computational and memory resources, a condition exacerbated by the recent trend towards extreme miniaturization, such as CubeSats and other types of micro and pico satellites \cite{furano_towards_2020}.
Energy is often the most limiting factor.
In fact, relatively high instantaneous power can be handled by power distribution units, and albeit still strict, computational bottlenecks are being removed as new dedicated processors like Tensor Processing Units (TPU) are tailored for use in space \cite{hinz_aivionic_2023}.
Still, available energy remains limited by engineering constraints (the maximum available mass and size of solar panels and secondary batteries) and operative conditions (i.e. eclipses).

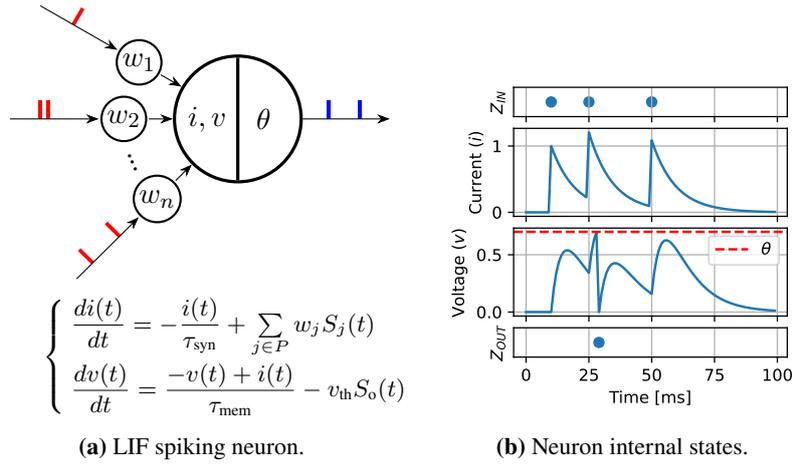
\begin{figure}[htbp]
    \centering
    \begin{subfigure}[b]{0.3\textwidth}
        \centering
        \begin{tikzpicture}[
        spike/.style={fill,thick,minimum width=0.05cm,minimum height=0.25cm,anchor=south,inner sep=0,outer sep=0},
        weight/.style={draw, circle, thick, inner sep=1pt, outer sep=1pt},
        figure/.style={draw=none, fill=none, inner sep=1pt, outer sep=1pt}]
        \node[figure,anchor=north] at (0,-2.3){
        \begin{footnotesize}
        $
        \begin{cases}
        \dfrac{di(t)}{dt} = -\dfrac{i(t)}{\tau_\text{syn}} + \sum\limits_{j \in P}w_j S_j(t)\\[2ex]
        \dfrac{dv(t)}{dt} = \dfrac{-v(t) + i(t)}{\tau_\text{mem}} - v_\text{th}S_\text{o}(t)
        \end{cases}
        $
        \end{footnotesize}
        };
        \node[draw, very thick, circle split, rotate=90] (N) at (0, 0) {\rotatebox{-90}{$i,v$} \nodepart{lower} \rotatebox{-90}{ $\theta$}};
        \coordinate [] (w2o) at (-3, 0);
        \node[weight] (w2) at (-1.5, 0) {$w_2$};
        \node (s11) at (-2.5,0) [spike,red] {};
        \node (s12) at (-2.6,0) [spike,red] {};
        \draw[-{Stealth}] (w2) -- (N);
        \draw[-{Stealth}] (w2o) -- (w2);
        \coordinate [] (w1o) at (-2.598, 1.5);
        \node[weight] (w1) at (-1.299, 0.75) {$w_1$};
        \node (s21) at (-2.165,1.25) [spike,red,rotate=-30] {};
        \draw[-{Stealth}] (w1) -- (N);
        \draw[-{Stealth}] (w1o) -- (w1);
        \node (dots) [rotate=-67.5] at (-1.386, -0.574) {...};
        \coordinate [] (wno) at (-2.121, -2.121);
        \node[weight] (wn) at (-1.061, -1.061) {$w_n$};
        \node (sn1) at (-1.909, -1.909) [spike,red,rotate=45] {};
        \node (sn2) at (-1.556, -1.556) [spike,red,rotate=45] {};
        \draw[-{Stealth}] (wn) -- (N);
        \draw[-{Stealth}] (wno) -- (wn);
        \coordinate [] (out) at (2.0, 0);
        \node (so1) at (1.2,0) [spike,blue] {};
        \node (so2) at (1.6,0) [spike,blue] {};
        \draw[-{Stealth}] (N) -- (out);
        \end{tikzpicture}
        \caption{LIF spiking neuron.}
        \label{fig:snn:neuron}
    \end{subfigure}
    \qquad
    \begin{subfigure}[b]{0.3\textwidth}
        \centering
        \includegraphics[width=\textwidth]{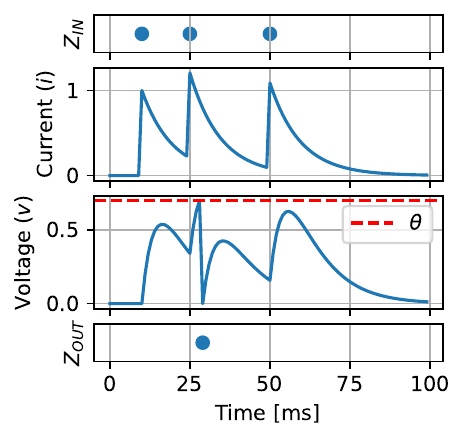}
        \caption{Neuron internal states.}
        \label{fig:snn:response}
    \end{subfigure}
    \caption{Current-based Leaky Integrate and Fire (LIF) neuron model. The neuron takes as input the pre-synaptic binary spikes $S_j$. The internal dynamic governs the trend of the current $i$ and voltage $v$. Whenever the voltage hits the threshold $\theta$, an output spike $S_o$ is emitted, and the voltage is reset to zero.}
    \label{fig:snn}
\end{figure}

In this context, Spiking Neural Networks (SNN) are highly attractive.
Inspired by the brain, where neurons communicate by means of discrete pulses (spikes), they promise a high computational efficiency. 
Although the internal dynamics of the specific neuron model determines its behavior \cite{izhikevich_which_2004}, the interface with the rest of the network operates exclusively by means of binary spikes.
All the neuron models share the same general behavior in which incoming spikes are accumulated in the internal states.
Whenever a certain condition is met (e.g. an internal state hitting a predefined threshold) an output spike is emitted.
In contrast to Artificial Neural Networks (ANN), computation is only performed at the time when a spike is received by a neuron, making the overall activity inside the network inherently sparse \cite{han_cross-layer_2018,kheradpisheh_temporal_2020}.
Moreover, computation between layers requires just Accumulate (AC) operations, since due to the binary nature of the spikes weights require to be just accumulated, in contrast with the Multiply and Accumulate (MAC) operations performed in classical ANN, further lightening the computation.
These features lower the demand for energy, making SNN highly attractive in contexts characterized by limited resources \cite{han_cross-layer_2018,bouvier_spiking_2019,kheradpisheh_temporal_2020,voelker_spike_2021,stockl_recognizing_2020,kheradpisheh_bs4nn_2022,hunsberger_training_2016,park_t2fsnn_2020,goltz_fast_2021}.
As an example, the dynamics and the response of a current-based Leaky Integrate and Fire (LIF) neuron, one of the most well-studied models, are shown in Fig.~\ref{fig:snn}.
As discrete spikes are essentially events, SNN are intrinsically compatible with event-based cameras, which have recently started attracting the interest of the space community for onboard applications such as navigation, landing, and visual tracking \cite{izzo2022neuromorphic}.
SNN can be implemented on event-based neuromorphic hardware, designed to fully exploit its asynchronous, sparse computation paradigm to achieve solutions capable of outperforming those based on ANN in terms of energy efficiency, potentially by multiple orders of magnitude \cite{han_cross-layer_2018,bouvier_spiking_2019,frenkel2023bottom}.
Nevertheless, the final performance is largely dependent on the combination of many parameters and design choices, such as the neuron model, the scheme of information encoding, target number of time steps at inference, and hardware implementation \cite{han_cross-layer_2018,bouvier_spiking_2019,kheradpisheh_temporal_2020,hunsberger_training_2016,comsa_temporal_2020}.
Despite such interesting features, neuromorphic research is a cutting-edge field, and at the time being there is no standard method for efficient training of SNN.
This is mainly due to the discontinuous nature of spikes, which hinders the use of well-known, gradient-based optimization algorithms, making state-of-the-art models struggle to scale to very deep architectures.
Methods based on temporal coding, also called latency-based methods, in which the information is carried in the time a neuron fires, offer the most promising trade-offs regarding energy optimization \cite{kheradpisheh_temporal_2020,goltz_fast_2021,stockl_recognizing_2020,kheradpisheh_bs4nn_2022,park_t2fsnn_2020,comsa_temporal_2020}. 
According to the Time-To-First-Spike (TTFS) coding, the more a neuron is activated, the shorter its firing delay, and the more significant the associated transmitted information. 
For classification tasks, Rank Order Coding (ROC) is even simpler, whereby the information is carried just in the firing sequence of the output neurons, regardless of the exact spike time.
In such encoding schemes, neurons can even be forced to fire once at most during inference, making SNN models based on temporal coding extremely attractive for energy-constrained applications.
In a benchmark comparison among different coding schemes, TTFS showed the highest performance in terms of accuracy, latency, power consumption and number of synaptic operations \cite{guo_neural_2021}. 
The energy efficiency benefits of time-coding based SNN for static data are shown in \cite{goltz_fast_2021} for the MNIST dataset on the BrainScaleS-2 neuromorphic processor. 

Overall, an effective demonstration of latency-based SNN on more complex tasks is still missing.
Proposed applications of SNN for space systems are still limited \cite{bose_spiking_2016,tao_novel_2005,sikorski2021event, lemaire_fpga-based_2020,mcleod2022globally,suresh_framework_2017,izzo2022neuromorphic}, and the ability of these models to cope with complex features and real-world tasks still awaits a convincing demonstration.
However, the functionality of the neuromorphic technology in the space environment has been recently proven with the successful testing of an Intel Loihi spiking processor on board of the TechEdSat-13 CubeSat \cite{murbach_techedsat-13_2022,ortiz_towards_2022,rahman_neuromorphic_2023}.

\subsection{Work objectives}
\label{sec:intro:objectives}
With neuromorphic research being a relatively new topic, several steps are still needed to move from theoretical and numerical studies to practical implementation on real hardware flying onboard spacecraft.
This work aims to contribute to this transition by providing some practical tools needed for the design of future spaceborne neuromorphic systems.
The primary objective is to estimate the actual advantages that can be expected from SNN with respect to classical ANN.
Here, the focus is put to the trade-off between accuracy and energy, as energy efficiency is the most prominent benefit sought in space applications.
To achieve this goal, a novel metric, capable of comparing the model complexity of both ANN and SNN in a hardware-agnostic way, is proposed as a proxy for the energy consumption.
The performance of several SNN and ANN models is compared  on a scene classification task, using the EuroSAT RGB dataset.
Special attention is placed on spiking models based on temporal coding, as they promise even greater efficiency of resulting networks, as they maximize the sparsity properties inherent in SNN, but rate-based models are included in the analysis as well.
In order to validate this approach, the energy trend predicted with the proposed metric is compared with actual measures of energy used by benchmark SNN models running on neuromorphic hardware.
The secondary goal of this work is to exploit the data collected in the comparison to analyze the internal dynamics of SNN models, identifying the most significant factors which affect the energy consumption, in order to derive design principles useful in future activities.

EuroSAT is a reference dataset for scene classification representative of a plausible use case in the Earth Observation field, namely land use classification \cite{helber_eurosat_2019}.
It includes multispectral and RGB imagery provided by the Sentinel-2A satellite, divided into 13 bands over 10 different land use and land cover classes.
The reason for this choice is twofold.
First, classification of satellite images is one of the tasks frequently considered as an application of onboard intelligence as it allows bandwidth requirements to be reduced by detecting and discarding invalid and corrupted images on board.
Second, ANN represent the current state of the art for this kind of task, which then constitutes the hardest possible benchmark to demonstrate the advantages of SNN.

The remainder of the paper is structured as follows.
The methodology followed in the formulation of test models and in the estimation of energy consumption is detailed in Sec.~\ref{sec:methodology}.
The results obtained by the numerical test campaign are analyzed in Sec.~\ref{sec:results}, and the capability of the proposed energy estimation method to predict energy consumption on hardware is presented in Sec.~{\ref{sec:hardware}}.
A discussion and a conclusion are given in Sec.~\ref{sec:conclusion}.

\section{Methodology}
\label{sec:methodology}
This section presents the followed methodology, including training methods, implemented models and methods for estimating the energy consumption.
Given the benchmark task, several models, both SNN and ANN, were trained to assess the impact of different architectures, neuron models, and encoding schemes on the energy consumption.

\subsection{Training method trade-off}
\label{sec:methodology:training}
The first trade-off consists in the selection of the training method for the spiking systems.
Looking for biologically plausible mechanisms, early researchers proposed unsupervised training methods like spike timing dependent plasticity (STDP), a form of Hebbian learning where weight changes depend on the relative timing between pre- and postsynaptic spikes \cite{caporale_spike_2008,diehl_unsupervised_2015,tavanaei_acquisition_2016}.
In STDP and other local learning rules, the training process does not depend on global error signals (as is the case with the training loss in the backpropagation algorithm), but weights are updated based only on information which is locally available to the weights and neurons themselves.
However, an unsupervised learning rule alone is not sufficient for decision making.
Some alternatives proposed to address this issue include coupling the training with reinforcement learning \cite{mozafari_bio-inspired_2019,mozafari_spyketorch_2019} or making use of local, layer-wise loss functions \cite{illing_local_2021}.
Nevertheless, STDP-based networks cannot be optimized for a specific target output, and they are yet to demonstrate their effectiveness on complex tasks.
Some approaches have been studied that allow rate-based conversion between deep ANN and SNN with minimal loss of accuracy \cite{stockl_recognizing_2020,han_cross-layer_2018,hunsberger_training_2016,diehl_fast-classifying_2015}.
However, the use of rate-based conversion usually requires high firing rates and a high number of time steps to provide acceptable performance \cite{stockl_recognizing_2020,neftci_surrogate_2019}. 
This approach seems to bring marginal benefits in terms of energy efficiency: the gain in energy savings seems to be reduced for complex datasets due to the higher number of time steps required, which increases energy consumption and leads to higher latency \cite{han_cross-layer_2018,bouvier_spiking_2019,kucik2021investigating,davidson2021comparison}. 
Alternative training methods were proposed to overcome the limitations imposed by spike discontinuity \cite{guo_direct_2023}.
Backpropagation of spike times \cite{kheradpisheh_temporal_2020,kheradpisheh_bs4nn_2022,park_t2fsnn_2020,comsa_temporal_2020,bohte_error-backpropagation_2002,mostafa_supervised_2018,sakemi_supervised_2021,klos2023smooth} leverages the fact that inter-spike time intervals are a continuous quantity.
Derivative chains of spike times are formulated analytically, so no approximation is involved, and backpropagation can be applied directly.
On the other hand, the method works only if strict constraints are imposed on both the neuron model and the network architecture, in order to enable the analytical formulation of the derivatives which in general are not available in closed-form solution, and once achieved, they are valid only for the specific model.
The Surrogate Gradient (SG) method has recently become a popular alternative as a training algorithm across the neuromorphic community \cite{zenke_superspike_2018,neftci_surrogate_2019,zenke_remarkable_2021,rossbroich_fluctuation-driven_2022,shrestha_slayer_2018}.
SG has the advantage of being agnostic to the neuron model and the type of encoding, with the ability to leverage traditional deep learning libraries, making it suitable for fast prototyping.
On the other hand, the unrolling in time requires large amounts of memory at training, which rapidly increases with the model size up to potentially impractical proportions, and it suffers from the problem of vanishing or exploding gradients.
Despite these potential drawbacks, SG was selected as the training method of choice for its ability to handle different test cases without the need to tailor the entire training for each of them.
In the following, unless explicitly stated otherwise, SG, in its SuperSpike \cite{zenke_superspike_2018} variant, was used as the training method for SNN, leveraging the Norse library \cite{pehle_norse_2021} for practical implementation.
To make the training compatible with latency-based coding, output spike trains are converted to spike times in a differentiable way, while maximal membrane voltage is used as decoding for rate-based networks \cite{eshraghian_training_2023}.

\subsection{Formulation of the test cases}
\label{sec:methodology:testcases}
The response of a SNN is determined by the dynamics of its neuron model \cite{izhikevich_which_2004}.
Biological plausibility was not included in the criteria considered in the formulation of the test cases; rather, energy efficiency was the main parameter of interest in the adoption of SNN in space applications.
Hence, only the most simple models were considered to minimize the amount of necessary computation.
In particular, two neuron models were considered: the leaky integrate-and-fire (LIF) and the non-leaky integrate-and-fire linear (IFL) neurons.
Three main types of network architectures were included in the test cases: Multilayer Perceptron (MLP), Convolutional Neural Network (CNN), and MLP with locally connected layers (MLP/LCL).
MLP/LCL is a MLP whose first layer takes 2D tensors (e.g., image data) as input while limiting the receptive field of each neuron in a way that mimics convolutional layers, which is one of the methods for circumventing limitations currently present on some neuromorphic processors which do not support convolution.
The effect of regularization on energy consumption for SNN was also analyzed.
Among the various possible methods, Batch Normalization Through Time (BNTT) was included in temporal coding-based models\cite{kim_revisiting_2021-1}, while layer-wise spike rate regularization, in which a target average fire rate is imposed for each layer, was used in rate-based networks.
More information about the dynamics of the adopted neuron models can be found in Table~\ref{tab:params-estimation}.
More information about network architectures is available in Appendix~\ref{app:networks}.

\subsection{Existing metrics for energy estimation}
\label{sec:methodology:metrics}
One of the aims of this work was the establishment of a rigorous way of comparing energy consumption among different Spiking Neural Networks, and between SNN and their ANN counterparts.
The intent was not to develop a method to estimate the absolute energy required, for the number of the unknown parameters is too high for such a task, but rather to achieve a credible way to perform relative comparisons, useful in evaluating trade-offs during the prototyping phase.
When dealing with ANN running on traditional computing hardware (CPU and GPU), data movement dominates energy consumption at inference, reaching up to \SI{90}{\percent} of the total for certain architectures \cite{yang_method_2017,horowitz_11_2014}.
Hence, even if widely used, the number of floating-point operations and the number of weights in the model might not be ideal metrics for energy computation.
In general, energy consumption can be expressed by the sum of two terms:
\begin{equation}
    E = E_\text{comp} + E_\text{mem}
\end{equation}
where $E_\text{comp}$ is the computational energy, and $E_\text{mem}$ is the memory energy consumed while moving data (weights and feature maps) to and from memory.
While the computational energy is directly proportional to the actual number and type of floating point operations, the estimation of the memory energy is more complex, for modern hardware organizes memory in a hierarchical manner, with different speed and cost of access depending on the application.
The way in which data flows across the hierarchy is one of the most prominent factors that affect and differentiate the various architectures (CPU, GPU with and without dedicated tensor operations, e.g., Nvidia tensor cores).
An accurate estimation of the energy would require a data flow model tailored for the specific network and layer architecture (i.e., layer types, dimensions, and sequence) \cite{yang_method_2017}.
It would seem reasonable to adopt a black-box approach as in \cite{degnan_assessing_2016} to estimate general metrics, expressed in \si{GMAC\per\watt} (billions of MAC operation per Watt) as an attempt to take into account the average contributions of both computational and memory energy consumption.
Nevertheless, this approach comes with its own limitations.
In recent years, hardware tailored for machine learning has seen an impressive improvement in both absolute performance and power efficiency (see, for example, \cite{nvidia_corporation_gp100_2016} compared to \cite{nvidia_corporation_nvidia_2021}), mostly due to specialized computing units (tensor cores, TPU, etc.).
But such performance figures are achieved by means of extreme parallelism, which entails the processing of data in \emph{batches}.
In fact, there is a large difference in energy consumption between batched and non-batched data in such type of hardware, especially due to the fact that most of the memory optimization strategies are less effective without massive parallelization \cite{davies_advancing_2021}.
Unfortunately, the estimation of the energy spent in non-batched data processing cannot be generalized and would require direct measurement on specific use cases.
Moreover, the energy efficiency drops quickly whenever the hardware capacity is not saturated, due to the relatively high power consumption in idle mode (a modern GPU can require up to \SI{80}{\watt} while idling\footnotemark{}).
The most used metrics to compare SNN performances are the number of emitted spikes and the number of synaptic operations \cite{guo_neural_2021,kim_revisiting_2021-1}.
Even if a certain increasing trend in energy consumption can be expected with increasing number of spikes, different internal connectivity can lead to a very different number of synaptic operations, making such a metric only a very rough estimation.
In \cite{guo_neural_2021}, the power consumption of different coding schemes in SNN is estimated and compared by implementing very simple benchmark networks on an FPGA. 
While useful for comparing different coding schemes, this method is hardly effective in comparing large networks and different neuron models since an equivalent HW implementation should be realized for each architecture.
Moreover, a comparison with equivalent ANN models would be impossible.
Some metrics have been developed to take into account the memory contribution to the energy consumption \cite{lemaire_analytical_2023,dampfhoffer_are_2023}.
Nevertheless, they still require some knowledge or assumption about specific hardware features, like the size of the I/O buffer in \cite{lemaire_analytical_2023} or the fraction of data reuse in \cite{dampfhoffer_are_2023}.

\footnotetext{Measurement taken on a Nvidia RTX A6000 GPU board mounted in a system with an Intel(R) Xeon(R) W-2275 CPU and 128 GB RAM. The reported power value is related only to the GPU.}

\subsection{Energy estimation procedure}
\label{sec:methodology:energy}
In this context, the number of \emph{Equivalent Multiply and Accumulate Operations (EMAC)} is proposed as a \emph{hardware-agnostic} method suitable for comparing the computational load of SNN based on different neuron models as well as for comparison with their ANN counterparts.
EMAC is intended as a dimensionless generalization of MAC operations, taking into account both computation and memory costs.
In fact, all the floating-point operations performed in standard ANN are of the Multiply and Accumulate (MAC) type.
Conversely, in SNN a significant part of the computation entails simpler Accumulate (AC) operations, thanks to the binary nature of spikes.
Then, different relative weights need to be assigned to AC and MAC in the estimation of the related computational burden.
\cite{horowitz_11_2014} reports a \SI{0.9}{\pico\joule} energy consumption for AC and \SI{4.6}{\pico\joule} for MAC operations for the FP32 number format in a \SI{45}{\nano\meter} CPU architecture, taking into account only the actual computation energy (excluding the memory contribution).
In this work, it is assumed that moving data in memory is the dominant factor in the determination of the energy performance: assuming a unitary weight for MAC ($\SI{1}{MAC} = \SI{1}{EMAC}$), this leads to a conservative weight value of $2/3$ assigned to the AC ($\SI{1}{AC} = \SI{0.667}{EMAC}$), due to the fact that in an AC operation only 2 numbers are involved, instead of 3 as in the MAC.
No assumption is made on the actual implementation of the networks.
Only the strict number of floating-point operations necessary for inference is included in the estimation, with no overhead related to input/output operations, system idle power consumption, or computation overhead due to actual implementation.
The assumptions above imply a further underlying assumption that the network is implemented in a digital fashion.
Although analog neuromorphic processors promise unprecedented improvements in energy consumption \cite{schemmel_accelerated_2020,banerjee_prospect_2022,indiveri_neuromorphic_2011}, a rigorous comparison between different networks would be even harder since the actual energy consumption is extremely dependent on the specific technological solution, often tailored to specific types of networks (i.e., neuron models).
The assumption of a digital implementation allows the achievement of more reliable comparison between different architectures.
For the sake of simplicity, in the following sections the terms \emph{energy consumption} and \emph{computational load} both refer to the value of EMAC required by the network to perform a single inference.    

In a SNN, part of the operations (such as the \emph{synaptic operations}) are performed whenever a spike is received, while other operations are executed at each time step during the update of the neurons internal states.
Then, the total number of EMAC operations $\text{E}_\text{tot}$ depends on two distinct contributions:
\begin{equation}
    \label{eq:energytot}
    \text{E}_\text{tot} = \text{E}_\text{syn} + \text{E}_\text{upd}
\end{equation}
where $\text{E}_\text{syn}$ is the synaptic operation contribution, and $\text{E}_\text{upd}$ is the neuron update.
The two terms can be further expanded in:
\begin{equation}
    \text{E}_\text{tot} = s e_\text{syn} + nT e_\text{upd}\label{eq:energy}
\end{equation}
where $s$ is the total number of synaptic operations performed during the inference, $n$ is the number of neurons in the network, and $T$ is the number of time steps.
The two terms $e_\text{syn}$ and $e_\text{upd}$ are respectively the energy per synaptic operation and the energy per neuron update, and they depend on the specific neuron model.
The general form of Eq.~\eqref{eq:energy} remains valid even for ANN, where it is treated as a special case in which there is no update in time and connections between layers act as synapses ($e_\text{upd}=0,\ e_\text{syn}=\SI{1}{EMAC}$).
The procedure to identify the energy parameters of a specific neuron model are as follows:
\begin{enumerate}
\item Write the neuron's dynamics in its discrete-time form (as in its digital implementation);
\item Identify each operation as MAC or AC;
\item Operations processed at each time instant are related to neuron update and contribute to $e_\text{upd}$;
\item Operations performed only at the time an input spike is received are synaptic operations and contribute to $e_\text{syn}$.
\end{enumerate}
The contribution of spike generation/voltage reset is neglected, as it can be considered just a variable assignment operation.
Table~\ref{tab:params-estimation} summarizes the estimation process for LIF and IFL neurons.

\begin{table}[htbp]
\centering
\caption{Energy parameter estimation for LIF and IFL neuron models. In the model descriptions, $i$ is the current, $v$ the potential, and $P$ denotes the domain of the presynaptic neurons.
$S_j(t) = \sum_k \delta(t-t_j^k)$ is the input spikes train from the $j$-th presynaptic neuron, where $\delta$ is the Dirac delta function and $k$ is the index of the spikes times.
$- v_\text{th}S_\text{o}(t)$, where $S_\text{o}(t)=\delta\bigl(v_\text{th}-v(t)\bigr)$ and $v_\text{th}$ is the threshold potential, represents the neuron reset mechanism.
}\label{tab:params-estimation}
\begin{footnotesize}
\begin{tabularx}{13.3cm}{|p{0.3cm}|p{6.cm}|X|}
    \hline
    &
\multicolumn{1}{c|}{LIF}
    & 
\multicolumn{1}{c|}{IFL}
    \\
    \hline
\rotatebox[origin=l]{90}{Continuous}
    &
\vbox{
\begin{equation}    
\begin{cases}
\dfrac{di(t)}{dt} = -\dfrac{i(t)}{\tau_\text{syn}} + \sum\limits_{j \in P}w_j S_j(t)\\[2ex]
\dfrac{dv(t)}{dt} = \dfrac{-v(t) + i(t)}{\tau_\text{mem}} - v_\text{th}S_\text{o}(t)
\end{cases}
\end{equation}   
}
     &  
\vbox{
\begin{equation}
\begin{cases}
\dfrac{di(t)}{dt} = \sum\limits_{j \in P}w_j S_j(t)\\
\dfrac{dv(t)}{dt} = i(t) - v_\text{th}S_\text{o}(t)
\end{cases}
\end{equation}
}
     \\
     \hline
\rotatebox[origin=l]{90}{\hspace{2.5em}Discrete}
    &
\vbox{
\begin{equation}
\begin{cases}
i_{k+1} = i_k - i_k\dfrac{\Delta t}{\tau_\text{syn}} + \sum\limits_{j=1}^{n_{S}}w_jS_{jk} + b \\[1ex]
v_{k+\frac{1}{2}} = v_k + (i_{k+1}-v_k)\dfrac{\Delta t}{\tau_\text{mem}} \\[1ex]
v_{k+1} = v_{k+\frac{1}{2}} - v_\text{th}S_{k+1}
\end{cases}
\end{equation}
}
     & 
\vbox{
\begin{align}
i\Delta t = \tilde{i},\ w_j\Delta t = \tilde{w}_j,\ b\Delta t = \tilde{b}, \nonumber\\[1ex]
\begin{cases}
\tilde{i}_{k+1} = \tilde{i}_k + \sum\limits_{j=1}^{n_{S}}\tilde{w}_jS_{jk} + \tilde{b} \\[1ex]
v_{k+\frac{1}{2}} = v_k + \tilde{i}_{k+1}\\[1ex]
v_{k+1} = v_{k+\frac{1}{2}} - v_\text{th}S_{k+1}
\end{cases}
\end{align}
}
    \\
    \hline
\rotatebox[origin=l]{90}{Parameters }
    &
\vbox{
\begin{equation}
\begin{array}{l}
e_\text{syn} = \SI{1}{AC} = \SI{0.667}{EMAC} \\
e_\text{upd} = \SI{2}{AC} + \SI{2}{MAC} = \SI{3.333}{EMAC}
\end{array}
\end{equation}
}
    &
\vbox{
\begin{equation}
\begin{array}{l}
e_\text{syn} = \SI{1}{AC} = \SI{0.667}{EMAC} \\
e_\text{upd} = \SI{2}{AC} = \SI{1.333}{EMAC}
\end{array}
\end{equation}
}
    \\
    \hline
\end{tabularx}
\end{footnotesize}
\end{table}

The estimation of the energy consumption can be then completed by the identification of the remaining parameters of Eq.~\eqref{eq:energy}.
Given the specific network architecture, the number of neurons $n$ is immediately known.
In case of TTFS/ROC encoding, the inference ends at the time the first spike is emitted by the output layer. That time instant is considered for energy estimation, leading to an even lower computational burden.
This implies also that the inference time $T$ in those cases is not fixed and can vary for each single inference: latency performance can still be assessed in terms of mean and standard deviation.
Conversely, for rate-based encoding the time of inference $T$ is constant and predefined.
The estimation of the number of synaptic operations $s$ can be performed layer-wise.
The number of synaptic operations performed between a layer $l$ and the previous layer $l-1$ can be estimated by:
\begin{equation}
    s_{(l)} = n_{s(l)} \, n_{n(l)} \, T \, P_{(l-1)}(S)\label{eq:nsynops}
\end{equation}
where $n_\text{s}$ is the number of pre-synaptic connections in the neuron receptive field and $n_\text{n}$ is the number of neurons in the layer, both of which are dependent on the layer type (convolutional, fully connected etc.).
$T$ is the number of time steps, and $P_{(l-1)}(S)$ is the probability that a pre-synaptic neuron emits a spike.
This last term can be estimated by:
\begin{equation}
P_{(l-1)}(S) \sim \dfrac{ N_{s(l-1)} }{T \, n_{n(l-1)}}\label{eq:spikeprob}
\end{equation}
in which $N_{s(l-1)}$ is the total number of spikes emitted by the previous layer at inference.
Substituting \eqref{eq:spikeprob} into \eqref{eq:nsynops}, we obtain:
\begin{equation}
s_{(l)} \sim n_{s(l)} \, n_{n(l)} \, T \, \dfrac{ N_{s(l-1)} }{T \, n_{n(l-1)}}
=
n_{s(l)} \, n_{n(l)} \, \dfrac{ N_{s(l-1)} }{n_{n(l-1)}}
=
n_{s(l)} \, n_{n(l)} \, f_{(l-1)}\label{eq:synops}
\end{equation}
where $f_{l-1}$ is the average spiking rate of the \emph{previous} layer.
Whenever applied to a ANN case, Eq.~\eqref{eq:synops} is still valid with $f_{(l-1)} = 1$.

Additional synaptic operations are performed inside \emph{recurrent layers}, if present.
In this case, the layer output spikes are routed recursively as input to the layer itself: depending on the layer type (convolutional, fully connected, etc.), such recursive routing is itself a convolution or a linear connection.
In this case, the number of synaptic operations due to the recursion $s_{r(l)}$ can be computed as:
\begin{equation}
s_{r(l)} = n_{s(l)} \, n_{n(l)} \, f_{(l)}\label{eq:synops-recurrent}
\end{equation}
which is similar to Eq.~\eqref{eq:synops}, except that the spiking rate \emph{of the same layer} $f_{(l)}$ is used.
The complete procedure followed to compute EMAC for a specified model is summarized in Algorithm~\ref{alg:emac}.

\begin{algorithm}
\caption{EMAC computation algorithm}\label{alg:emac}
\begin{algorithmic}
\State Create network
\If{$T$ is constant} \Comment{Constant latency per inference, e.g. rate based networks}
  \State set $T$
\EndIf
\For{each layer $l$}
  \State compute $e_{\text{syn}(l)}$, $e_{\text{upd}(l)}$ \Comment{Neuron model can be different for each layer}
  \State compute $n_{s(l)}$, $n_{n(l)}$
  \State $n_{sr(l)} \gets 0$ \Comment{Default non-recurrent layer}
  \If{$l$ is a recurrent layer}
    \State compute $n_{sr(l)}$ \Comment{Number of recurrent synapses per neuron}
  \EndIf
\EndFor
\State Train the model
\For{each sample in Test Set}
  \State Perform inference on the sample
  \If{$T$ is variable} \Comment{e.g., in case of TTFS/ROC encoding}
    \State compute $T$ \Comment{As the time step the first spike is emitted at output}
  \EndIf
  \For{each layer $l$}
    \State compute $f_{(l)}$
  \EndFor
  \State $E_\text{tot} \gets \sum_l \underbrace{ \Bigl( n_{s(l)}n_{n(l)}f_{(l-1)} + n_{sr(l)}n_{n(l)}f_{(l)} \Bigr) }_{s_{(l)}} e_{\text{syn}(l)} + T\sum_ln_{n(l)}e_{\text{upd}(l)}$ \Comment{Compute energy}
\EndFor
\State Compute energy statistics over the Test Set (mean, standard deviation).
\end{algorithmic}
\end{algorithm}

\subsection{Known limitations}
\label{sec:methodology:limitations}
The energy estimation approach here presented has both strengths and weaknesses.
Some clear advantages are achieved with respect to other metrics usually adopted, like counts of weights, floating-point operations, or spikes.
The most obvious is the possibility to compare very different models, specifically SNN and ANN.
At the same time, the separation of synaptic operations and neuron updates allows to take into account the impact of the model size and latency, but also of its internal connectivity, all independent factors indistinguishable by other methods.
In analyzing the results, it must be taken into account that some network architectures are not equally efficient when implemented on different hardware.
In fact, while ANN are nowadays extremely efficient on hardware like GPU and TPU (if batched computation is applicable), their implementation on neuromorphic hardware could be almost impossible.
The opposite is valid for SNN: while on neuromorphic devices they promise to be extremely efficient in terms of energy, with no need to operate in batch mode, their implementation on standard hardware is complex: the use of libraries such as PyTorch make prototyping and training SNN very straightforward yet slow and memory-consuming since the network has to be unrolled in time.
Moreover, while the network is trained offline, it could still be possible to process input in batches. However, in a real-world application this is often impossible.
In this regard, SNN on neuromorphic hardware would have a clear advantage.
Another issue that makes the estimation of absolute energy consumption even more hardware-dependent is the fact that on standard computing devices (CPU/GPU/TPU) the relative weight of a floating point operation is always the same, while it is not necessarily the same on neuromorphic hardware, where, even in digital implementation, synaptic operations and neuron update may happen in different parts of the processor, leading to different energy consumption even for operations that are formally of the same type \cite{davies_loihi_2018}.
Still, the approach followed in this work remains useful in this context.
Being hardware-agnostic, it allows relative comparisons between architectures, enabling at least a preliminary information for trade-off.
The possibility to distinguish contributions due to synaptic operations with respect to neuron updates provides useful insight to better understand and optimize SNN models during preliminary design phases.
It is worth noting that the approach followed here is intentionally conservative in treating SNN, with the aim to present an upper bound on the energy estimation in spiking systems.
First, the number of time steps in SNN is not optimized at training, with a significant impact on the $E_\text{upd}$ term in Eq.~\eqref{eq:energytot}.
Second, the energy model does not take into account data locality, assigning the relative weights of MAC/AC operations based only on the number of quantities involved in the computation.
Normally, the more local the involved data are stored with respect to the computing cores, the lower the energy cost.
This property would tend to further favor SNN once implemented on neuromorphic processors, where weights are stored in a highly parallel fashion, local to the single neuron (or to small groups of neurons).
I SNN can achieve a better dimensionless energy consumption compared to a similar ANN, it can then be expected that such advantage would be even greater once ported on neuromorphic hardware.
The proposed method does not take into account any energy consumption contribution but the operations strictly necessary for inference.
In particular, idle power consumption and possible overhead due to general API implementation are not taken into account: these two terms can have a very large variability depending on the actual hardware platform and should be considered in a practical trade-off estimation.
Finally, the specific numerical representation of the mathematical operations is intentionally not considered.
While it is true that energy consumption can be dramatically different by reducing the bit depth, or by changing from floating to fixed point arithmetic, model quantization can be performed in both ANN and SNN alike.
Nevertheless, the proposed metric can be easily adapted to take into account different formats, if needed, simply by changing the relative weights attributed to MAC/AC operations accordingly.

\subsection{Section summary}
\label{sec:methodology:summary}
\begin{itemize}
\item 
Surrogate Gradient (SG) is selected as training method for SNN since it is not dependent on any specific neuron model nor encoding, enabling the training of several different networks in a simple way;
\item
in the formulation of the test cases, LIF (rate coding, spike rate regularization) and IFL (spiking once at most, ROC temporal coding, BNTT regularization) are selected as spiking neuron models;
\item
both MLP and CNN are considered as main architectures, in both SNN and ANN versions;
\item
existing metrics usually considered in the literature to estimate energy consumption in SNN are not suitable for the type of analysis constituting the main goal of this work;
\item
Equivalent MAC operations (EMAC) per inference is proposed as a hardware-agnostic proxy for the energy consumption.
This novel metric differentiates the contribution of synaptic operations and neuron update, can compare SNN and ANN, and it can take into account the impact of any neuron model, under the assumption of a digital implementation. A conservative approach is adopted to ensure that overestimation of the SNN benefits is avoided.
\end{itemize}

\section{Results}
\label{sec:results}
Following the methodology outlined above, 82 test models are trained.
The test cases are different combinations of neuron models and network architectures.
In contrast with other training methods (i.e. conversion from ANN), the use of Surrogate Gradient imposes the selected coding scheme (rate or temporal based) \emph{only at the output layer}, allowing the behavior of intermediate layers of the network to be governed by backpropagation.
Nevertheless, for simplicity of notation, the networks are here referred as rate or temporal based, even if the different schemes are applied just at the output decoding.
This use of terminology is generally accepted \cite{eshraghian_training_2023}.
Three neuron models are considered:
a) LIF spiking neurons with rate encoding (RATE);
b) IFL spiking neurons with Rank Order Coding (ROC), allowed to spike once at most during the inference;
c) non-spiking ReLu neurons (ANN).
Four general architectures are included:
A) Multilayer Perceptrons (MLP)
B) MLP architecture with Locally Connected Layer (MLP/LCL);
C) Convolutional Neural Networks (CNN);
D) MLP with a single recurrent hidden layer (MLP/REC)
All the model were trained with SG.
The naming convention of the test cases is summarized in Table~\ref{tab:case-label}.
It should be noted that several hyperparameters (referred to both the model and the training process) must be taken into account to determine each specific case.
To avoid impractically long labels, only the main ones are synthesized in the naming scheme.

\begin{table}[htb]
\centering
\caption{Test case naming convention.}\label{tab:case-label}
\begin{small}
\begin{tabularx}{\textwidth}{p{3.5cm}X}
\hline
\multicolumn{2}{c}{ID\_ + Type + Architecture + Nlayers + [SpecialProperty]} \\
\hline
ID & Unique numeric ID \\
Type & \textbf{R} (rate coded SNN) $|$ \textbf{T} (rank order coded SNN) $|$ \textbf{A} (ANN) \\
Architecture & \textbf{M} (Multilayer Perceptron) $|$ \textbf{C} (Convolutional) \\
Nlayers & Number of network layers \\
SpecialProperty (opt.) & \textbf{L} (Locally connected Layer) $|$ \textbf{R} (Recurrent) $|$ \textbf{B} (BNTT reg.) $|$ \textbf{F} (Spike rate reg.) \\
\hline
\end{tabularx}
\end{small}
\end{table}

\subsection{Accuracy vs. energy}
\label{sec:results:energy}
\begin{figure}[htb]
  \centering
  \includegraphics[width=\textwidth]{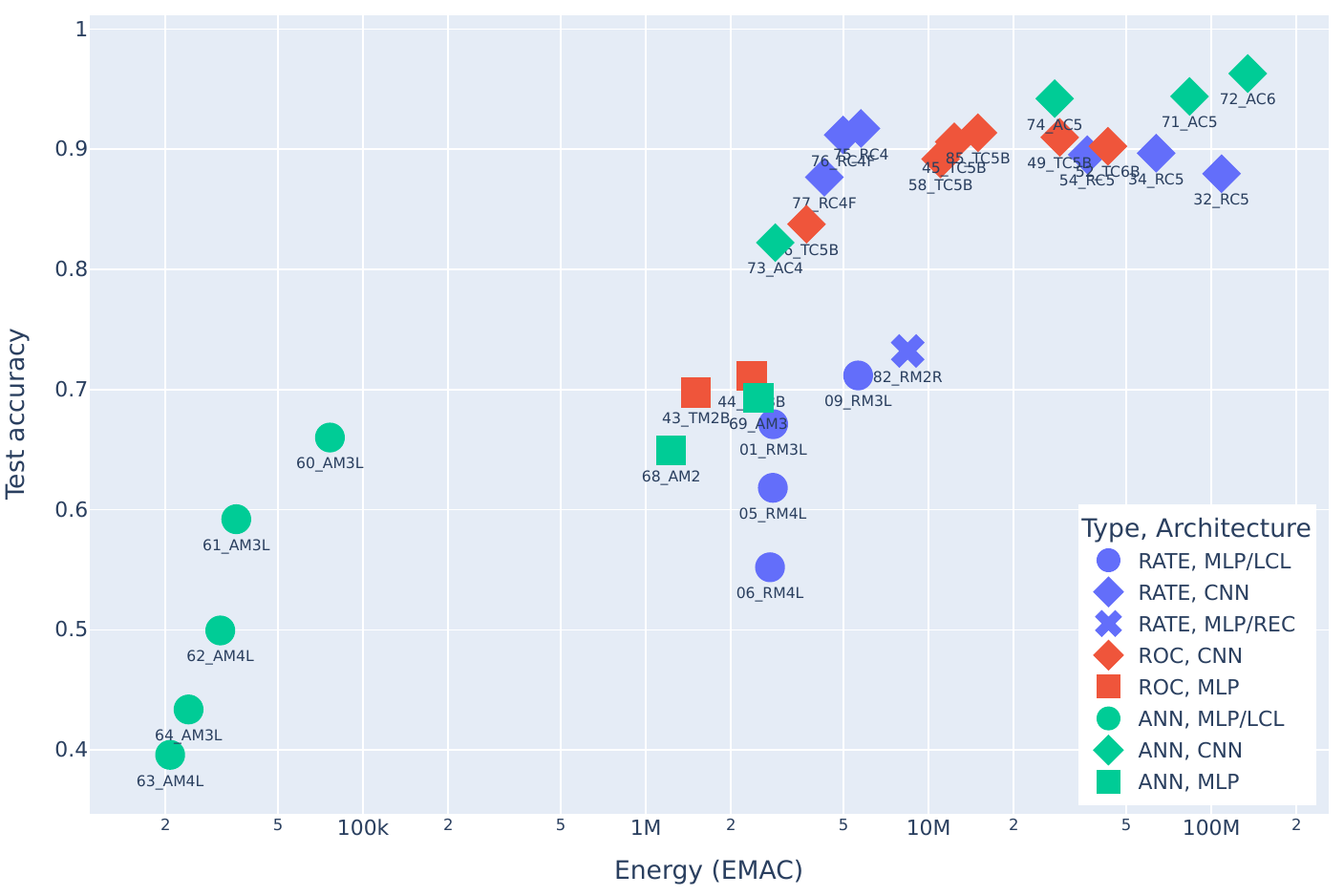}
  \caption{Accuracy vs. dimensionless energy (EMAC).}\label{fig:accuracy_vs_energy}
\end{figure}

The first and most important figure of merit is the achieved accuracy with respect to the energy consumption.
Fig.~\ref{fig:accuracy_vs_energy} shows the obtained results.
Here, we show only the results of tests in which the highest performance was achieved for each of the groups mentioned, to highlight what could be considered a Pareto front of the achievable performance.

The first, and most notable, result is that SNN are able to reach performance comparable of ANN, with significantly lower energy consumption.
This is particularly evident looking at the latency architectures \emph{45\_TC5B}, \emph{49\_TC5B}, and \emph{52\_TC6B}, corresponding respectively to the ANN cases \emph{74\_AC5}, \emph{71\_AC5}, and \emph{72\_AC6}: a $\sim\SI{2.5}{\percent}$ drop in absolute accuracy corresponds to an average $\sim\SI{60}{\percent}$ decrease in the EMAC per inference.
The same trend is observed for rate-based networks trained from scratch with SG, where architecture \emph{75\_RC4} achieves better performance than latency-based ones on both accuracy and EMAC.
Overall, spiking CNN exhibit a drop in EMAC per inference in the range \SI{-50}{\percent} to \SI{-80}{\percent} with respect to their ANN counterparts.
This result is in agreement with those reported in \cite{lemaire_analytical_2023} and \cite{dampfhoffer_are_2023}.
The opposite can be observed for simpler networks: both spiking MLP and spiking MLP with Locally Connected Layer show increased performance with respect to their ANN counterparts in term of accuracy, highlighting the ability of spiking system to store a large amount of information efficiently.
Nevertheless, for such simple networks, this is achieved at the expense of a higher EMAC value.
Moreover, the accuracy of spiking MLP and MLP/LCL networks remains relatively low (under \SI{75}{\percent}), a value which is not competitive with convolutional architectures, albeit better than their ANN counterparts.
It is noteworthy that a prominent role in the determination of the total EMAC per inference in SNN is played by the number of time steps in the simulation, a hyperparameter whose optimization has not been specifically pursued in this work.
Its role is even more important in simple networks in which the reduced number of neurons forces the information to be compressed, reducing in this way the sparsity of the computation (with a high number of synaptic operations due to the fully connected architecture of the MLP).
In this case, a non optimized number of time steps acts as a multiplier in energy consumption.
The Locally Connected Layer, applicable only to the first layer in the network to handle this specific problem, seems insufficient to significantly limit the impact of the subsequent fully connected layers.
Conversely, the inherently small receptive field of the convolutional layers is particularly well suited to exploiting the sparsity of spiking computation.

\begin{figure}[htb]
  \centering
  \includegraphics[width=0.65\textwidth]{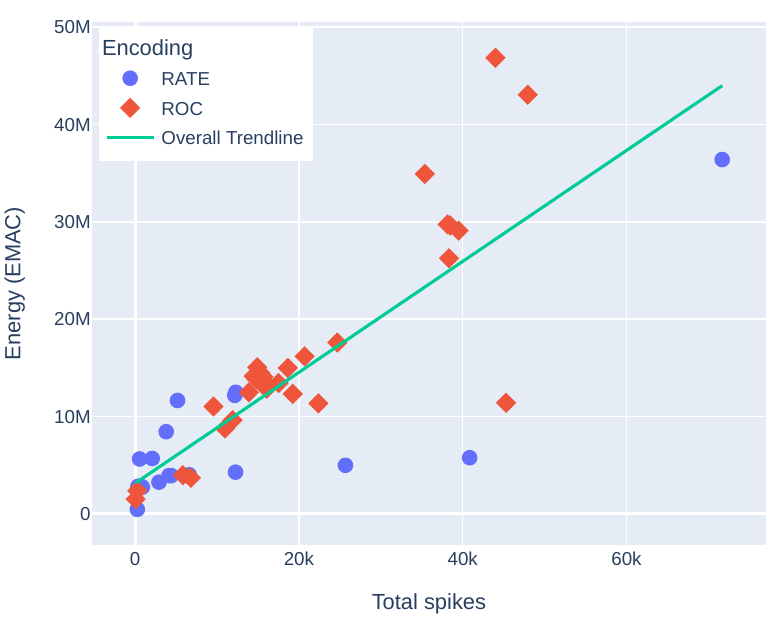}
  \caption{Dimensionless energy (EMAC per inference) w.r.t. total emitted spikes per inference, average values over the test set.}\label{fig:energy_vs_spikes}
\end{figure}

\begin{figure}[htb]
  \centering
  \includegraphics[width=0.75\textwidth]{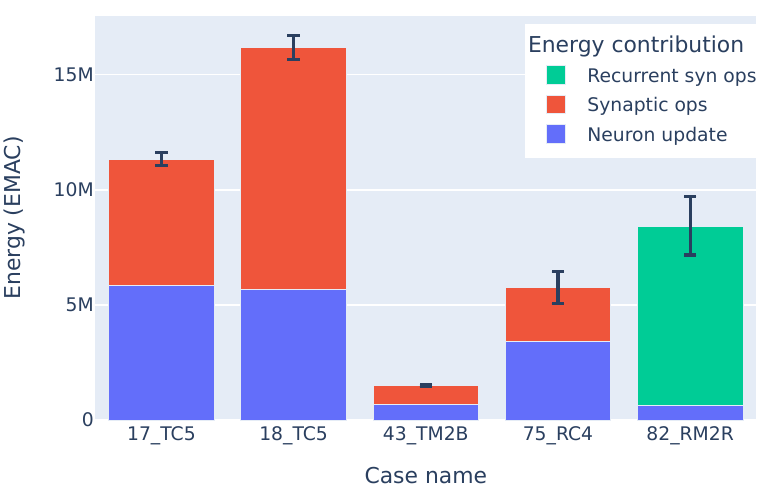}
  \caption{EMAC per inference for selected test cases: breakdown across network tasks, average value over the test set. Standard deviation value is indicated by black error bars.}\label{fig:energy_breakdown}
\end{figure}

\begin{figure}[htb]
  \centering
  \includegraphics[width=0.75\textwidth]{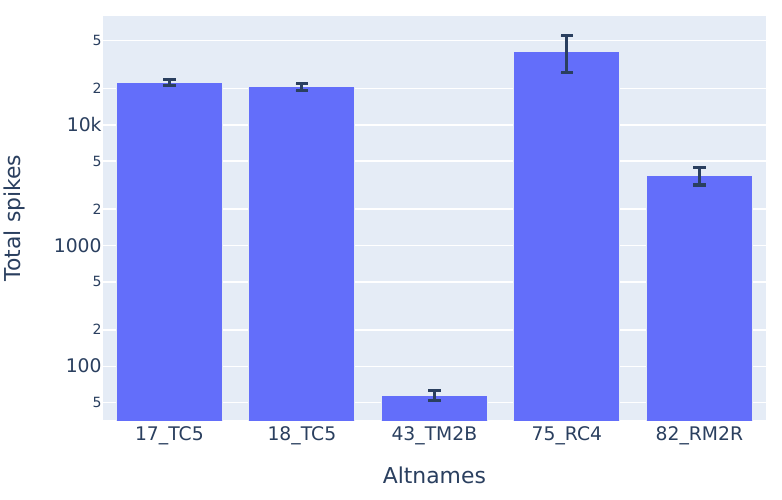}
  \caption{Total emitted spikes per inference, average value over the test set. The standard deviation value is indicated by black error bars.}\label{fig:spikes_breakdown}
\end{figure}

Fig.~\ref{fig:energy_vs_spikes} compares the proposed metric EMAC/inf with the most simple metric often adopted in literature as first estimation, that is the number of emitted spikes at inference.
The average values evaluated on the test set are used to build the chart.
Even if a general trend is present (as the spikes increase, the EMAC/inf are generally larger) it can be seen how the use of the number of spikes as the only metric could lead to large errors with a large variation of computational load among cases with almost the same number of spikes, and vice versa.
The reason of this behavior resides in the fact that a significant part of the energy consumption is related to synaptic operations.
Hence, the actual routing of spikes between layers, and the network's internal connectivity pattern can drastically change the overall energy consumption, even with a similar number of emitted spikes.
Fig.~\ref{fig:energy_breakdown} shows the breakdown of EMAC/inf among neuron update, synaptic operations, and recurrent synaptic operations for some selected test cases, while Fig.~\ref{fig:spikes_breakdown} shows their respective number of emitted spikes.
Histograms show the average value, while error bars shows the standard deviation.
Cases \emph{17\_TC5} and \emph{18\_TC5} share the same convolutional architecture, in terms of neuron types, latency encoder/decoder, and number and size of spiking layers.
The only difference between the two is the connection pattern between convolutional layers: in \emph{17\_TC5}, convolution is performed with kernel size $k=3$ and stride $s=1$, and dimensional reduction is achieved with standard maxpool layers; conversely, in \emph{18\_TC5} convolution is performed with $k=5$ and $s=2$, with no maxpooling.
Due to the larger receptive field of neurons, \emph{18\_TC5} exhibits a $\sim\SI{45}{\percent}$ increase in energy, even with $\sim\SI{10}{\percent}$ less spikes.
Looking at the energy breakdown, sharing the same number and type of neurons, and with an almost equal average number of time steps at inference (\SI{38.1}{\milli\second} and \SI{37.1}{\milli\second}) \emph{17\_TC5} and \emph{18\_TC5} require a practically equal amount of EMAC for neuron update. In this case, the increase is instead mostly due to a large difference in the energy spent for synaptic operations.
From an energy consumption perspective, it would appear advisable to favor a low degree of internal connectivity.
However, it must be said that a larger receptive field in the convolution seems to bring significant benefit to the system performance, with \emph{18\_TC5} scoring a $+\SI{5.3}{\percent}$ in the absolute test accuracy with respect to \emph{17\_TC5}.
A careful trade-off analysis between energy and accuracy should be performed in the determination of an effective architecture.

The discrepancy between the number of spikes and the computational load is even more evident in the case of \emph{75\_RC4}, which is still a convolutional architecture.
Being a rate-based network, the number of emitted spikes is high, and almost doubles the values achieved by \emph{17\_TC5} and \emph{18\_TC5}.
Nevertheless, the lower number of time steps (\SI{32}{\milli\second}), and the lower size and number of layers reduce the average energy to almost a third of that of \emph{18\_TC5}.
\emph{43\_TM2B} and \emph{82\_RM2R} are another notable example showing the importance of internal connectivity.
They both share the same MLP architecture with one hidden layer of 100 neurons, but while \emph{43\_TM2B} is a spiking MLP with ROC encoding, the hidden layer of \emph{82\_RM2R} is made by recurrent LIF neurons with rate encoding.
From Fig.~\ref{fig:energy_breakdown} is possible to see how the computational effort required at inference due to neuron update is practically the same in the two cases.
But recurrence involves a higher number of synaptic operations, leading to a total value of EMAC per inference more than 5 times higher, despite the higher efficiency in storing information due to the recurrent neurons, which reflects in a significant $+\SI{3}{\percent}$ absolute increment in test accuracy.
In fact, in case \emph{82\_RM2R}, $>\SI{90}{\percent}$ of the computational effort is due to synaptic operations in the recurrent layer.

\subsection{Effects of regularization}
\label{sec:results:regularization}
\begin{figure}[htb]
  \centering
  \includegraphics[width=0.6\textwidth]{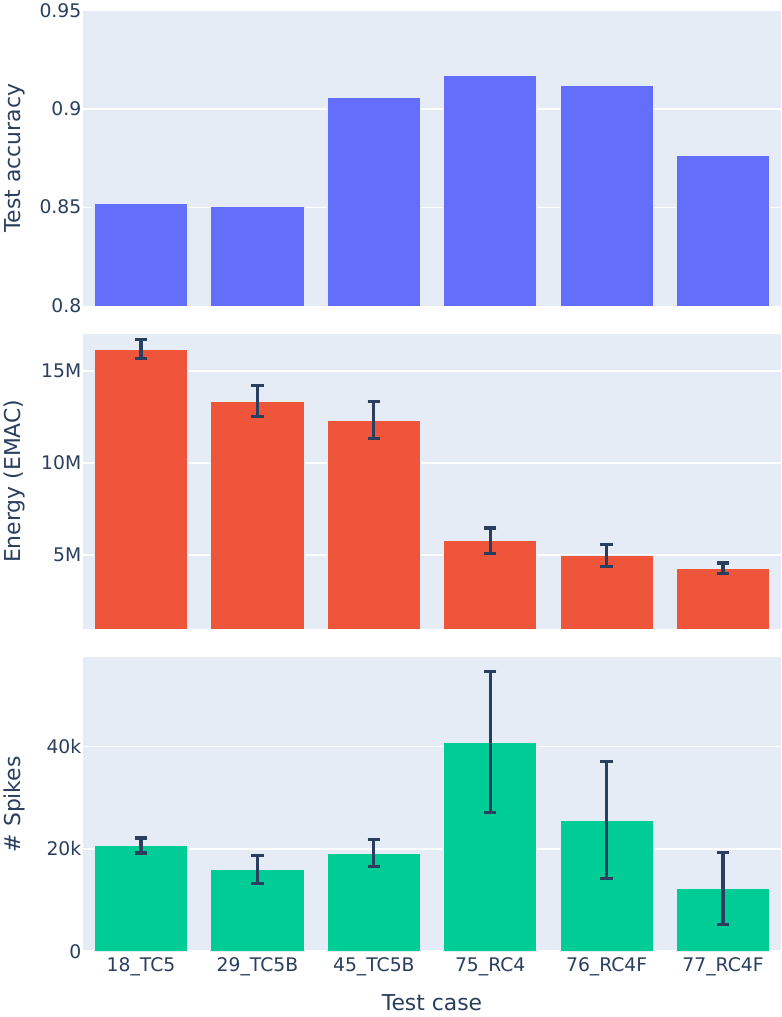}
  \caption{Effect of regularization. \emph{xx\_TC5x} cases make use of Rank Order Coding and BNTT; \emph{7x\_RC4x} cases are rate-based with target spiking rate normalization.}\label{fig:bntt}
\end{figure}

Fig.~\ref{fig:bntt} shows the impact of different forms of regularization in the training of SNN from scratch.
The histograms report test accuracy, EMAC/inference, and the number of emitted spikes per inference for 3 test cases with Rank Order Coding adopting BNTT (\emph{18\_TC5}, \emph{9\_TC5B}, and \emph{45\_TC5B}), and 3 models based on rate encoding (\emph{75\_RC4}, \emph{76\_RC4F}, \emph{77\_RC4F}).
For each regularization method, the models share the same architecture, with different regularization settings.
In the ROC/BNTT cases, \emph{18\_TC5} entails no regularization; neuron-wise BNTT is applied in \emph{9\_TC5B}; spatial BNTT is applied in \emph{45\_TC5B}.
The number of spikes is similar in all three cases (with a lightly reduced value for \emph{9\_TC5B}).
Neuron-wise BNTT seems beneficial for the energy consumption, but with no improvement on the test accuracy.
Spatial BNTT improves both the accuracy, with a \SI{5.4}{\percent} gain in the absolute test accuracy, and the energy consumption, with a \SI{24}{\percent} drop in EMAC per inference compared to case \emph{18\_TC5}.
Rate limiting regularization seems to achieve an opposite result: no regularization is applied in \emph{75\_RC4}; in \emph{76\_RC4F} a target average spiking rate of 2 spikes per neuron per inference was set; the same value was set to 1.5 in \emph{77\_RC4F}.
In these cases, more restrictive regularization provokes a decrease in accuracy, with a limited improvement in energy consumption.
This limited gain can be explained by the internal distribution of the energy consumption.
In fact, as visible in Fig.~\ref{fig:energy_breakdown}, the energy consumption in the \emph{75\_RC4} architecture is dominated by neuron update, while limiting the spiking rate impacts only the contribution by synaptic operations.
A better results could be achieved by conjugating rate-limiting regularization with other methods designed to reduce the network latency.

\begin{figure}[htb]
  \centering
  \includegraphics[width=0.6\textwidth]{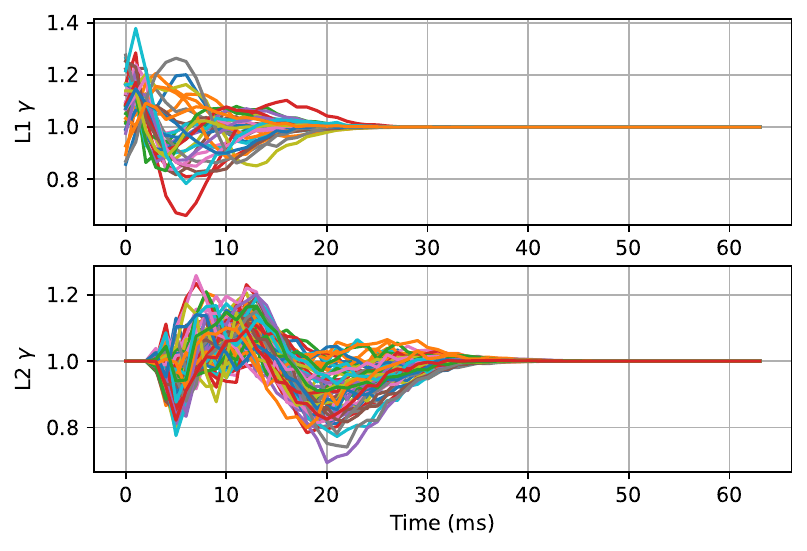}
  \caption{\emph{45\_TC5B}, Batch Normalization Through Time, learned $\gamma$ in the two convolutional layers. The network is capable to identify a temporal pattern in the incoming spike trains.}\label{fig:bntt_gammas}
\end{figure}

BNTT can also be exploited to derive some additional insight on the internal functioning of the network.
Fig.~\ref{fig:bntt_gammas} shows the values of the output scale parameter $\gamma$ for the two convolutional layers in case \emph{45\_TC5B}.
In each layer, there is a value of $\gamma$ for each time step, for each convolution channel, since spatial BNTT is applied.
In this case, $\gamma$ was initialized at 1 at the beginning of the training.
It is clearly visible how a temporal pattern is correctly learned for each channel.
However, in both layers it is possible to also determine the time interval during which the values of the parameters has changed during training: from zero to \SI{25}{\milli\second} for the first layer, from \SI{3}{\milli\second} to \SI{40}{\milli\second} for the second one.
This range is shifted forward in time for the higher layer, as could be expected, as it takes a certain amount of time for the potential in the lower layer to increase enough to spike.
This also shows the time period in which each layer affects the network output, e.g. the fact that backpropagation has left the values of $\gamma$ in layer 2 unchanged after \SI{40}{\milli\second} is an indication that any spike emitted by the layer after that time has a zero or negligible impact on the final output.
This information could be exploited in future developments to further optimize the training process.

\subsection{Scaling to deeper architectures}
\label{sec:results:scaling}
\begin{figure}[htb]
  \centering
  \includegraphics[width=0.75\textwidth]{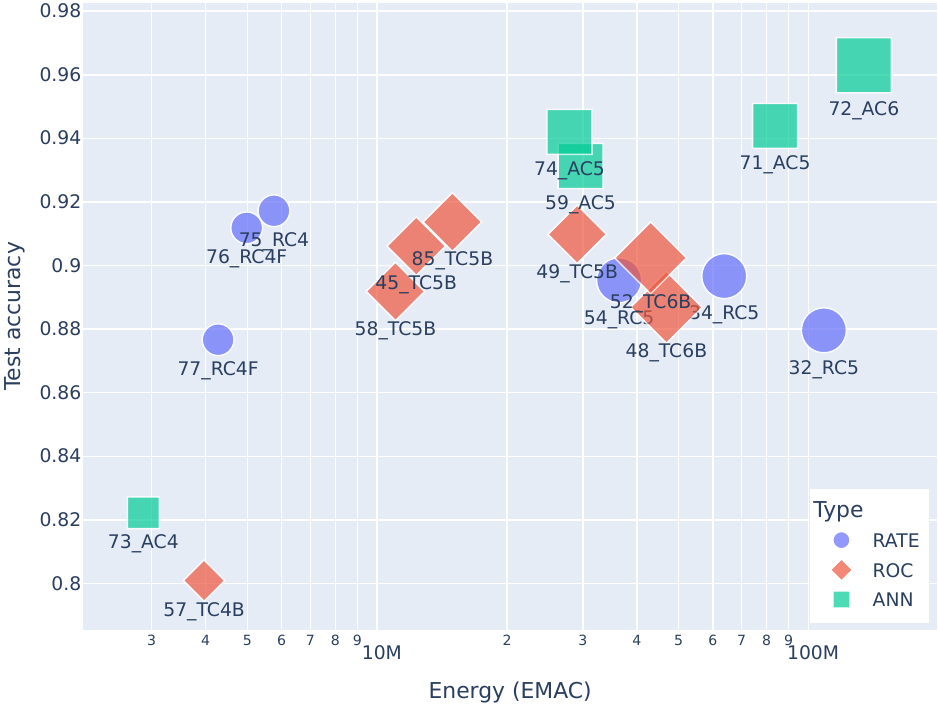}
  \caption{SNN scalability. The size of the markers is related to the number of layers (ranging from 3 to 6).}\label{fig:scalability}
\end{figure}

Despite the regularization schemes tested in this work, and the generally promising performance levels achieved in terms of accuracy at inference, it is still a struggle to scale SNN to deeper architectures.
In this section, the behavior of SNN with respect to layer scaling is analyzed.
Fig.~\ref{fig:scalability} is a close-up of Fig.~\ref{fig:accuracy_vs_energy}, limited to the convolutional architectures.
Differently from Fig.~\ref{fig:accuracy_vs_energy}, not just the top performing models are shown, but also some other test cases designed to study the performances achieved when equal architectures (in term of layer type and sequence) are applied to different types of network (latency-based SNN, rate-based SNN, ANN).
All the models in the figure are CNN with number of layers varying from 3 to 6 (denoted by the marker size in the scatter plot).
Latency-based (ROC), rate-based (RATE) and their ANN counterparts are included.

ANN show a consistent behavior in which as the number of layers increases, the accuracy improves.
Preliminary tests have shown that VGG-style ANN of moderate depth ($\sim10$ layers) are capable of achieving accuracy \SI{>98}{\percent}.
SNN exhibit a different trend: while at lower complexities they outperform ANN, showing the ability of spiking neurons to store information, as the network depth increases, they reach a sort of maximum: at this point they seem to preserve a good performance, with a clear advantage in terms of energy (as EMAC per inference) with respect to their ANN counterparts.
However, as their structural complexity increases further, the performance in terms of accuracy starts to drop to unsatisfactory levels.

\begin{figure}[htbp]
     \centering
     \begin{subfigure}[b]{0.42\textwidth}
         \centering
         \includegraphics[width=\textwidth]{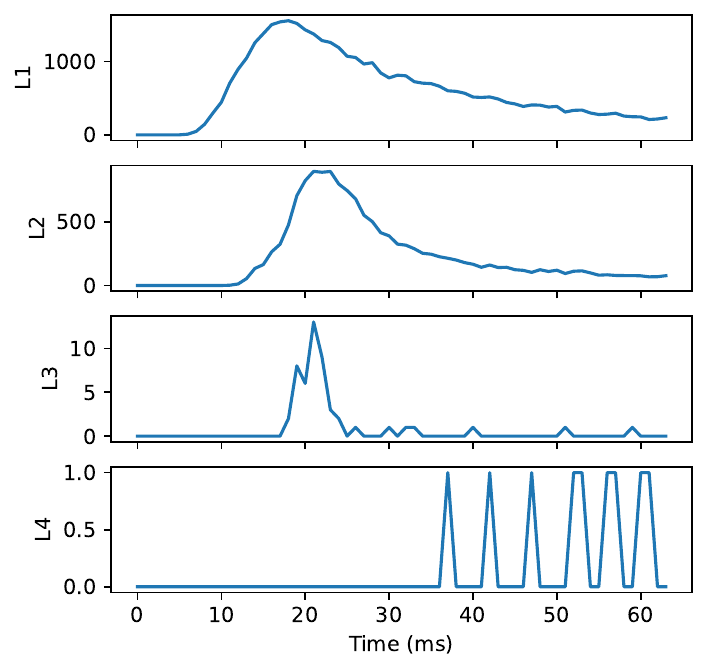}
         \caption{Case 49\_TC5B -- effective training.}
         \label{fig:spikes1}
     \end{subfigure}
     \begin{subfigure}[b]{0.42\textwidth}
         \centering
         \includegraphics[width=\textwidth]{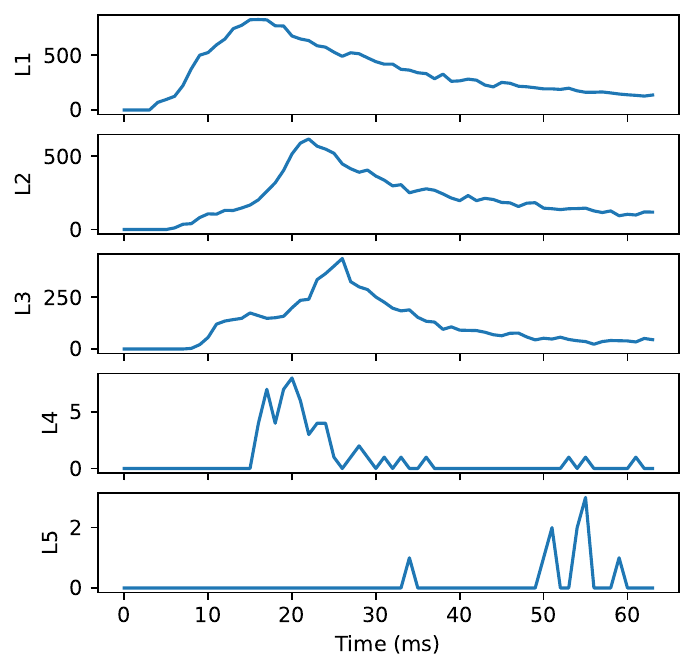}
         \caption{Case 52\_TC6B -- suboptimal training.}
         \label{fig:spikes2}
     \end{subfigure}
     \\
     \begin{subfigure}[b]{0.42\textwidth}
         \centering
         \includegraphics[width=\textwidth]{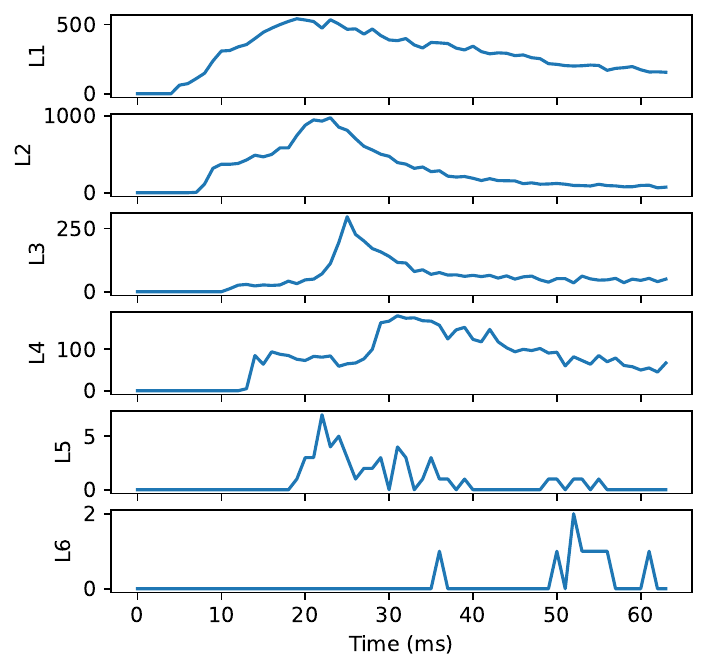}
         \caption{Case 46\_TC7B -- suboptimal training.}
         \label{fig:spikes3}
     \end{subfigure}
     \begin{subfigure}[b]{0.42\textwidth}
         \centering
         \includegraphics[width=\textwidth]{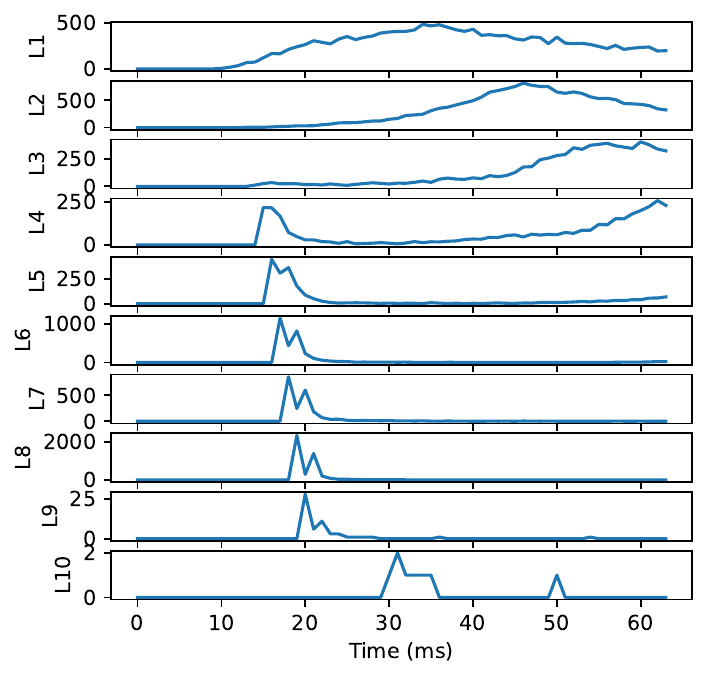}
         \caption{Case 25\_TC11B -- failed training.}
         \label{fig:spikes4}
     \end{subfigure}
        \caption{Scaling to deeper networks with spiking CNN and Rank Order Coding (IFL neurons). In the four examples above, BNTT is applied on convolutional layers (all except the last two layers, which are fully connected). As the number of layers grows, subsequent layers fail to collect most of the information of lower layers, and their firing is based only on a few presynaptic spikes.}
        \label{fig:spikes_all}
\end{figure}

A deeper insight of this phenomenon can be obtained by looking at the internal dynamics of the networks.
Fig.~\ref{fig:spikes_all} shows the temporal trend of the number of spikes emitted by each layer of different models with increasing number of layers.
All the models in the figure are latency based: \emph{49\_TC5B}, \emph{52\_TC6B}, and \emph{46\_TC7B} are also included in Fig.~\ref{fig:scalability}; \emph{25\_TC11B} is a 11-layer model (10 plus the convolutional spiking encoder) of the same network type that failed training (with a final test accuracy equal to \SI{14.26}{\percent}, slightly higher w.r.t. a random guess).
\emph{49\_TC5B} in Fig.~\ref{fig:spikes1} represents a case of a well-trained SNN: in the two convolutional levels, the layer takes a certain amount of time to collect spikes from its predecessor, and the neurons' potential increases.
As the voltage levels approach the threshold, the number of emitted spikes starts to increase, reaches a maximum, and then slowly decreases, as less significant neurons spike.
As can be expected, the response of subsequent layers is slightly translated in time, as each layer needs to accumulate a certain amount of spikes emitted by the previous layer.
In case \emph{52\_TC6B} (Fig.~\ref{fig:spikes2}), with just one convolutional layer more, the trained response is already suboptimal.
In fact, at the second layer the rise in the number of spikes proceeds in two steps: first, around \SI{6}{\milli\second} the spike level rises, stabilizing at \SI{\sim 9}{\milli\second}; a second, more relevant rise starts at \SI{15}{\milli\second}, culminating in the layer's peak at \SI{\sim 20}{\milli\second}.
At layer 3, this behavior is even more evident, with a secondary peak at \SI{\sim 12}{\milli\second} and the main peak at \SI{25}{\milli\second}; this leads the subsequent layer 4 to its peak of activity \emph{well before the previous layer has emitted its most relevant spikes}.
Although synchronicity is recovered between layers 4 and 5 (the output layer), that means that the final output \emph{depends on partial information}.

As the number of layers increases, this behavior is even more pronounced.
In case \emph{46\_TC7B} in Fig.~\ref{fig:spikes3}, layer 5 reaches its top activity before the peak is reached by the two previous layers, 4 and 3.
As the depth continues to increase, this phenomenon compromises the training.
Case \emph{25\_TC11B} in Fig.~\ref{fig:spikes4} is a clear example: spike activity in layer 4 rises and reaches its peak in response of a very few input spike from the previous layer.
In this way, most of the information included in the input is not involved in the inference, resulting in a poor final accuracy.
A possible explanation of this phenomenon is that the number of time steps at training was too low for the deeper networks.
With the same weight initialization, as the number of layers increases, the time needed by the spikes to propagate across subsequent layers up to the output in the early epochs becomes more large, possibly larger than the available time (which is a hyperparameter).
This means that backpropagation starts when only a subset of the input is available, and several neurons do not significantly contribute to the training.
Nevertheless, the network latency at training is limited by the large amount of memory required, due to the network unrolling in time.
Memory consumption at training time scales linearly with the size of the network (with a constant number of time steps) and linearly with the number of time steps;
however, even if the final latency can be tuned after training (i.e., with TTFS/ROC coding, in which the inference ends as the first output spike is emitted), deeper networks require a larger number of time steps at training in order to allow a proper propagation of the spikes in the network at the beginning of the process. Thus, in practice, the memory usage at training scales similarly to the square of the network size.
Moreover, this could also limit the effectiveness of some regularization techniques, such as BNTT.
In fact, enabling training with limited memory can be achieved by reducing the batch size, which on the other hand can degrade the precision of statistical quantities like mean and variance of the input adopted in the regularization.
Also, a higher number of time steps implies an accumulation over time of intrinsic gradient estimation errors due to surrogate gradient formulation, potentially hindering the training even further.
A more detailed investigation is needed to unlock successful training of deeper networks.
Two possible directions for future investigations are possible: network initialization and regularization during training.

\subsection{Section summary}
\label{sec:results:summary}
\begin{itemize}
\item
Convolutional SNN are capable of reaching performance comparable to ANN, with significantly lower (\SI{-50}{\percent} to \SI{-80}{\percent}) EMAC per inference;
\item
such performance is achieved by both temporal and rate coding when networks are trained from scratch as spiking models (i.e., not converted from ANN);
\item
internal connectivity plays a crucial role in determining the global energy consumption, significantly affecting the contribution related to synaptic operations;
\item
spatial BNTT applied to SNN with temporal coding benefits both the accuracy and the energy consumption, while spike rate regularization associated with rate coding has a detrimental effect on the accuracy, with a limited benefit on the energy consumption;
\item
SNN still struggle to scale to deeper architectures: when the number of layers $N\geq5$, higher layers start to fire while only partial information is available from lower layers. Possible causes for this include the large memory consumption at training and error accumulation, both due to the unrolling in time adopted by SG, which can limit the effectiveness of regularization methods such as BNTT.
\end{itemize}

\section{Hardware testing}
\label{sec:hardware}
\begin{figure}[htb]
    \centering
    \includegraphics[width=0.6\textwidth]{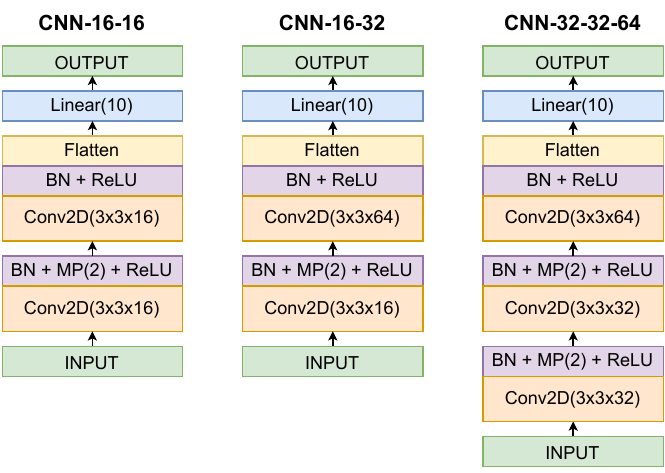}
    \caption{Models trained for the Akida device. ANN models, prior to the conversion in SNN, are shown.}\label{fig:akida-models}
\end{figure}

In order to evaluate the energy consumption of spiking networks on actual hardware, a series of benchmark models were implemented on the BrainChip Akida AKD1000 \cite{BrainChip_Akida_WP} device as it has built-in power consumption reporting capabilities.
The AKD1000 system-on-chip (SoC) is the first generation of digital neuromorphic accelerator by BrainChip, designed to facilitate the evaluation of models for different types of tasks, such as image classification and online on-chip learning.
Three convolutional models compatible with the AKD1000 hardware were trained on the EuroSAT dataset.
They consists of convolutional blocks made up by a Conv2D $\rightarrow$ BatchNorm $\rightarrow$ MaxPool $\rightarrow$ ReLU $\rightarrow$ stack of layers.
The last convolutional block lacks the MaxPool layer and is followed by a flattening and a dense linear layer at the end.
The three models differ in the number of convolutional blocks, and in the number of filters in each Conv2D layers: the architectures are detailed in Fig.~\ref{fig:akida-models}.
The selection of rather small models was determined by the hardware capabilities, for a single AKD1000 node was available during this work.
The Akida architecture can be arranged in multiple parallel nodes to host larger models.
An overview of the model components and parameters is given in Table~\ref{tab:akida_models_params}.

\begin{table}[ht]
    \centering
    \caption{Summary of the training parameters for the Keras model and the quantized model.}
    \label{tab:akida_models_params}
    \scriptsize
    \begin{tabular}{|c c | c c|}
        \hline
         \multicolumn{2}{|c|}{\bf Keras model} & \multicolumn{2}{c|}{\bf Quantized model} \\
         \hline
         Conv2D & kernel size: ${3}\times{3}$ & Input/weight/activation bits & $8/4/4$ \\
         MaxPool & pool size: ${2}\times{2}$ & Calibration epochs & $1$ \\
         ReLU & max. value: $6$ & Per-tensor activations & Enabled \\
         Optimiser & Adam & Optimiser & Adam \\
         Learning rate (start/end) & $10^{-3}$ / $10^{-5}$ & Learning rate (start/end) & $10^{-4}$ / $10^{-6}$ \\
         Weight decay & $4\times{10^{-4}}$ & Weight decay & $4\times{10^{-4}}$ \\
         Training epochs & $30$ & Training epochs & $20$ \\
         \hline
    \end{tabular}
\end{table}

\subsection{Training pipeline and dataset}
The Akida system is digital, and its predefined way to operate consist in first training standard ANN and then convert them into SNN to be run on hardware.
The current pipeline involves training a Keras model using the BrainChip MetaTF SDK, which builds upon the standard TensorFlow framework.
The trained Keras model is then converted to a quantized model with the aid of quantization tools built into the SDK, whereby it is possible to specify the number of bits per weight ($1-8$) depending on the types of layers.
The resulting quantized model inevitably suffers a drop in accuracy, which could be quite drastic depending on the structure and complexity of the model (e.g., from 90\% to 30\%).
This can be mitigated by applying calibration during quantization (the process of mapping the distribution of inputs to the quantization range determined by the number of bits used in the layer) and retraining.
The last step is the conversion of the quantized model in a spiking model that can run on the hardware.
Currently available documentation does not offer a clear explanation about the adopted neuron model (albeit it must be simple, for the AKD1000 does not support leaky neurons \cite{BrainChip_MetaTF}, and the implementation suggest that the converted spiking model reduces the computation to a single time step.
Nevertheless, the use of ROC decoding and the integrate-and-fire behavior in the Akida processor is confirmed in \cite{brainchip_what_2022}.

\subsection{HW Performances}
The Keras models managed to achieve test a accuracy value \SI{\sim80}{\percent} (Table~\ref{tab:akida_models_results}).
While the accuracy of the quantized model dropped due to the quantization process, during retraining it managed to recover accuracy values to levels very close to those of the Keras model.
This is attributed to the additional training time -- the quantized model was trained for an additional $20$ epochs -- and the fact that the training of the quantized model started from a relatively good (or at least \textit{non-random}) state.
The last conversion step (from the quantized model to the Akida spiking model executed on the AKD1000 hardware) did not have a perceptible effect on the accuracy.
In fact, for the larger model (\textit{CNN-32-32-64}), the accuracy of the Akida model even \textit{increased} by about \SI{1}{\percent} compared to the quantized model that it was created from.
The source of this increase is as yet unclear.
Overall, the final Akida spiking models suffered a drop in accuracy between \SI{1.3}{\percent} (\textit{CNN-16-32)} and \SI{5.6}{\percent} (\textit{CNN-16-16}), which is acceptable considering the drastic drop in bit depth (internal network parameters used only $4$ bits to represent values).
The average inference time ranges from \SI{0.66}{\milli\second} (CNN-16-16) to \SI{1.41}{\milli\second} (CNN-32-32-64), depending on the input size and on the depth of the model.

In terms of energy efficiency, the AKD1000 reported an extremely low energy consumption between \num{0.63} and \SI{1.38}{\milli\joule} per frame during inference.
The Akida offers also an estimation of the number of MAC operations executed at inference, albeit the built-in reporting functionality in the MetaTF software framework is still experimental and there is some uncertainties in the given values.
Nevertheless, such information has not been considered for a direct comparison with EMAC, for such a metric in its non-dimensional version does not offer an actual count of MAC operation, but just a proxy to compare the relative computational weights of different models.
It is worth to be noted that the reported value for the floor (idle) power consumption was \SI{911.49}{\milli\watt}, to be compared with a peak power consumption (worst case across the models) of \SI{978.00}{\milli\watt}.
This highlights the low power requested at the runtime, but on the other hand it must be considered that more than \SI{93}{\percent} of the overall power consumption is due to the device just being idle.
Overall, while the Akida AKD1000 demonstrated its potential for enabling efficient on-chip classification that could potentially be suitable for onboard space applications, it would also be necessary to take into account the floor power in any evaluation of a spacecraft power budget, especially in the case of nano and pico satellites (i.e. CubeSats).

\begin{table}[ht]
    \centering
    \footnotesize
    \begin{threeparttable}
        \caption{Summary of the network performances for the Keras quantized models.}
        \label{tab:akida_models_results}
        \begin{tabular}{l l l l}
             \hline
             & \textbf{CNN-16-16} & \textbf{CNN-16-32} & \textbf{CNN-32-32-64} \\
             \hline
             Test accuracy (Keras) 	&	\SI{79.33}{\percent}	&	 \SI{79.41}{\percent}	&	 \SI{85.56}{\percent}	\\
             Test accuracy (Quantized) 	&	\SI{74.19}{\percent}	&	 \SI{78.41}{\percent}	&	 \SI{80.85}{\percent}	\\
             Test accuracy (Akida) 	&	\SI{73.93}{\percent}	&	 \SI{78.11}{\percent}	&	\SI{81.22}{\percent}	\\
             Floor power [mW] 	&	911.49	&	911.49	&	911.49	\\
             Inference power (min--max) [mW] 	&	933.00--947.00	&	 933.00--949.59	&	 956.00--978.00 	\\
             Inference power (avg. $\pm$ sd) [mW] 	&	944.58 $\pm$ 3.86	&	 953.00 $\pm$ 4.42	&	 974.83 $\pm$ 4.06 	\\
             Inference energy [mJ/frame] 	&	0.63	&	0.82	&	1.38	\\
             Avg framerate [fps] 	&	1506.7	&	1154.8	&	707.55	\\
             \hline
        \end{tabular}
    \end{threeparttable}
\end{table}

\subsection{Hardware energy model identification}
The dimensionless EMAC metric can be adapted to estimate the actual energy consumption for a specific target platform, by conveniently tuning its internal parameters.
In this section the possibility to identify the AKD1000 energy model is investigated, using the following procedure:
\begin{enumerate}
    \item The 3 models implemented in the AKD1000 chip are reproduced in Norse, and their EMAC is computed;
    \item Two of them iare used to perform a least squares estimation considering $e_\text{syn}$ and $e_\text{upd}$ in Eq.~\eqref{eq:energy} as free parameters to fit the EMAC value to the actual energy consumption estimated by the hardware;
    \item The estimated model is used to compute the energy consumption of the third model, which is compared with the value measured on the Akida platform to assess the achieved accuracy.
\end{enumerate}
The more precisely the networks are reproduced in Norse, the higher the accuracy that can be expected from the energy model.
In the specific case, since some aspect of the conversion from ANN are undocumented, some arbitrary assumptions needed to be made in the Norse implementation.
Nevertheless, considering how the EMAC metric is modeled, such uncertainties are at least partially handled by proper tuning of the $e_\text{upd}$ parameter.
IFL neurons which spike once at most, with 64 time steps and Rank Order Coding has been selected.
To reproduce a realistic network activity, the Norse models have been retrained from scratch with SG, for the weights values trained in the MetaTF pipeline cannot be directly applied to the models.

\begin{figure}[htb]
  \centering
  \includegraphics[width=0.75\textwidth]{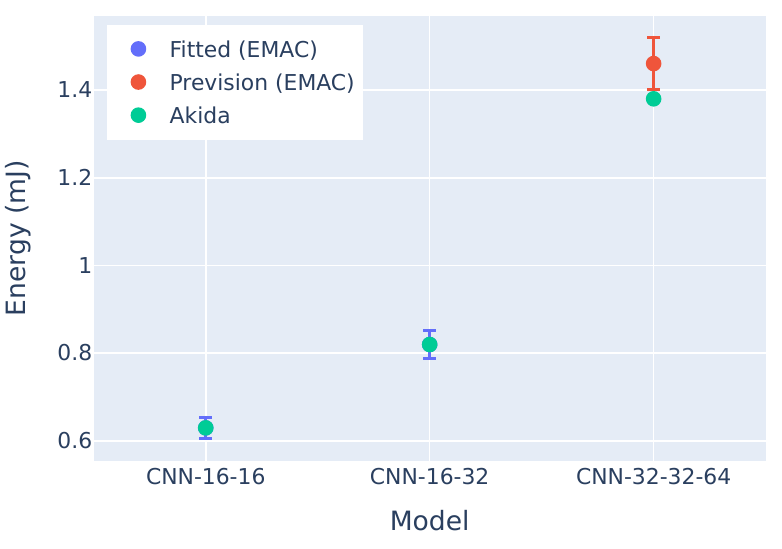}
  \caption{Identification of the Akida energy model. Vertical bars identify the $1\sigma$ uncertainty of the identified model.}\label{fig:identification}
\end{figure}

Results are shown in Fig.~\ref{fig:identification}.
Cases \textit{CNN-16-16} and \textit{CNN-16-32} have been considered in the identification of the numerical energy model.
As can be expected, by using the minimum number of data points (2 for a 2 DoF model) the residuals of the least squares optimization are practically zero and the estimated energy perfectly matches the measure for the two spiking models.
The obtained EMAC model has then been used to estimate the energy for \textit{CNN-32-32-64}, achieving a $+\SI{5.8}{\percent}$ relative error, just outside the $1\sigma$ uncertainty range of the model: the scaling of the energy consumption due to the increase of the network size is successfully described by the now dimensional EMAC metric.
It worth to be noted that the selection of only two data points for the estimation highlights the generalization capabilities of the proposed model.
Given more data, an even better accuracy in the estimation can be expected.

\subsection{Section summary}
\label{sec:hardware:summary}
\begin{itemize}
\item
EMAC parameters can be tailored to specific hardware to achieve absolute energy estimation;
\item
such functionality is tested on the BrainChip Akida AKD1000 neuromorphic processor;
\item
the measure of energy consumption of 2 SNN implemented on HW is used to identify EMAC parameters; the energy model is then used to estimate the HW energy consumption of a third network, achieving a \SI{5.8}{\percent} error;
\item
a minimal approach is intentionally adopted: additional measures are expected to improve the accuracy.
\end{itemize}

\section{Conclusion}
\label{sec:conclusion}
An investigation of the potential benefits of Spiking Neural Networks for onboard AI applications in space was carried out in this work.
The EuroSAT RGB dataset, a classification task representative of a class of tasks of potential interest in the field of Earth Observation, was selected as case study.
SNN models based on both temporal and rate coding, and their ANN counterparts, were compared in a hardware-agnostic way in terms of accuracy and complexity by means of a novel metric, Equivalent MAC operations (EMAC) per inference.
EMAC is suitable for assessing the impact of different neuron models, distinguishing the contributions of synaptic operations with respect to neuron updates, and comparing SNN with their ANN counterparts.
In its base formulation, EMAC achieves dimensionless estimation, and it should then be considered only a proxy for energy consumption.
Nevertheless, internal parameters can be tuned to match the features of specific hardware, if known, achieving also absolute estimation.
A preliminary successful demonstration is given for the BrainChip Akida AKD1000 neuromorphic processor.
Benchmark SNN models, both latency and rate based, exhibited a minimal loss in accuracy, compared with their equivalent ANN, with significantly lower (from \SI{-50}{\percent} to \SI{-80}{\percent}) EMAC per inference.
An even greater energy reduction can be expected with SNN implemented on actual neuromorphic devices, with respect to standard ANN running on traditional hardware.
While Surrogate Gradient proved to be an easy and effective way to achieve offline, supervised training of SNN, scaling to very deep architectures to achieve state-of-the-art performance is still an issue.
Further research is needed particularly in the search of architectures capable to exploit SNN peculiarities, and in the development of regularization techniques and initialization methods suited to latency-based networks.
Attention should be also given to recent developments in training techniques which do not require backward propagation in time, but only along the network at each time step \cite{yin_accurate_2023}, and new ANN-to-SNN conversion methods tailored for achieving extremely low latency \cite{yan_hypersnn_2023}.
Overall, the confirmed superior energy efficiency of Spiking Neural Networks is of extreme interest for applications limited in terms of power and energy, which is typical of the space environment, and SNN are a competitive candidate for achieving autonomy in space systems.

\section*{Ethics declarations}
This work was founded by the European Space Agency (contract number: 4000135881/21/NL/GLC/my) in the framework of the Ariadna research program.
The authors declare that they have no known competing financial interests or personal relationships that are relevant to the content of this article.
The EuroSAT dataset used in this activity is publicly available at \cite{helber_eurosat_2019}.
\appendix
\section{Benchmark network models}
\label{app:networks}
Adopted SNN models do not differ with respect to regular ANN, except for the activation functions, replaced by spiking neurons.
Multilayer Perceptrons (MLP) and Convolutional Neural Networks (CNN) are the building blocks used in this activity to define the test cases.
A MLP with Locally Connected Layer has also been developed to circumvent some possible limitations of certain neuromorphic hardware in implementing convolution.

\begin{figure}[htb]
  \centering
  \includegraphics[width=0.6\textwidth]{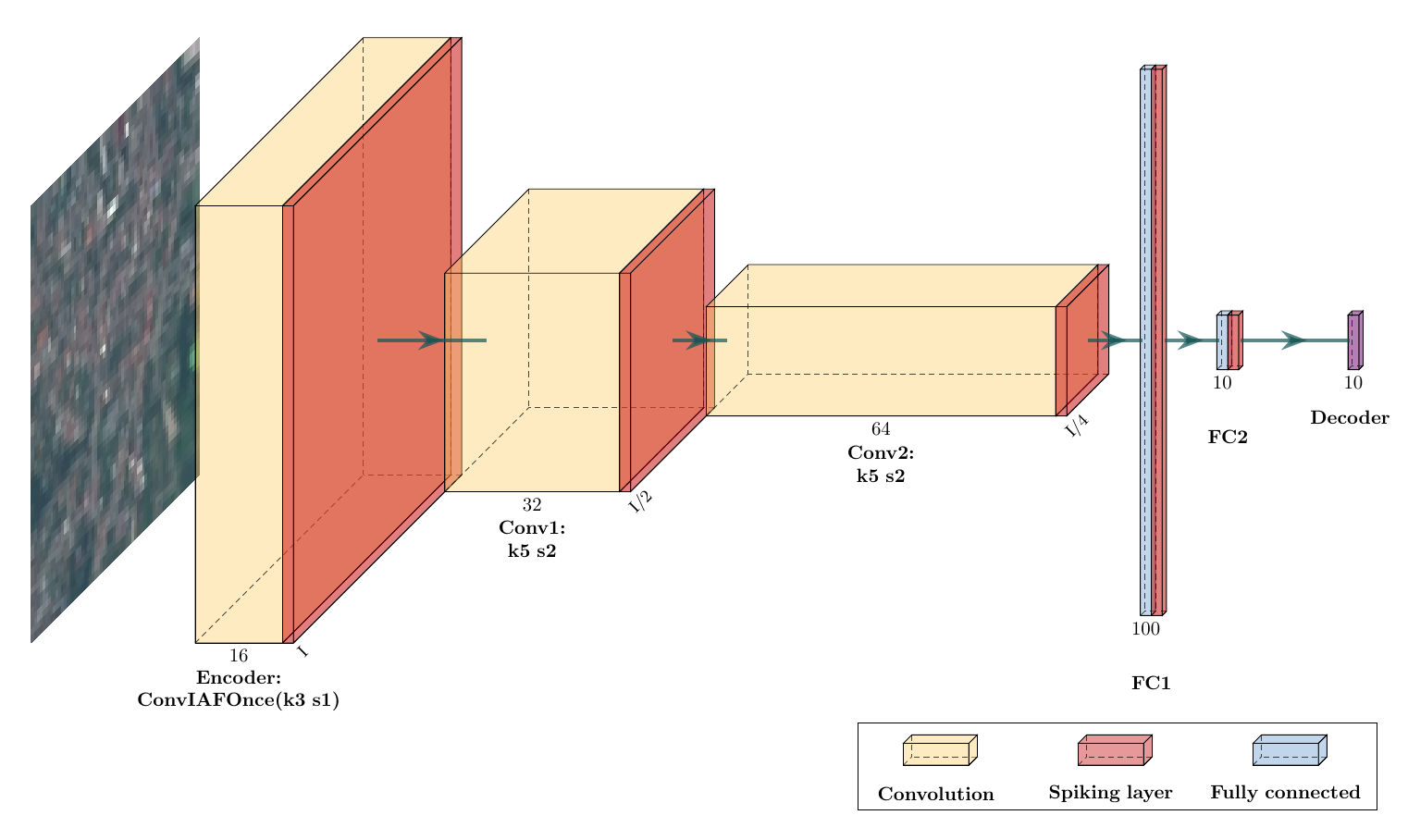}
  \caption{Benchmark convolutional architecture.}\label{fig:scnn_baseline}
\end{figure}

In the formulation of the test cases, the use of a common baseline architecture is useful to highlight the relative impact of single changes.
This allows to easily compare SNN-specific variations: neuron model, encoder/decoder models, neuron hyperparameters, and so on.
The most simple ANN architecture, Spiking MLP up to 3 hidden layer in size was included in the tests.
Regarding convolutional networks, a VGG-style general architecture was designed: it consists in a learnable convolutional encoder, which convolves the input and translate it in a spikes train, followed by two convolutional spiking layers.
Except at the encoder, at each convolution spatial dimensions halve in size, while the number of filters is doubled.
Two fully connected layers (a hidden layer and the output layer) are placed on top of the network.
The reduction in size along the spatial dimensions (height and width) is achieved by means of a stride value $s=2$ in the convolution, a solution that proved to be more efficient at training SNN in preliminary tests.
Nevertheless, also models with classic max pooling has been tested.
The baseline architecture, shown in Fig.~\ref{fig:scnn_baseline}, entails 3 convolutional layers (encoder included) and \num{114798} spiking neurons.
Variations with different number of layers, different neurons, types of regularization, size of layers, and encoding/decoding styles have been studied.

\begin{figure}[hbt]
  \centering
  \includegraphics[width=0.5\textwidth]{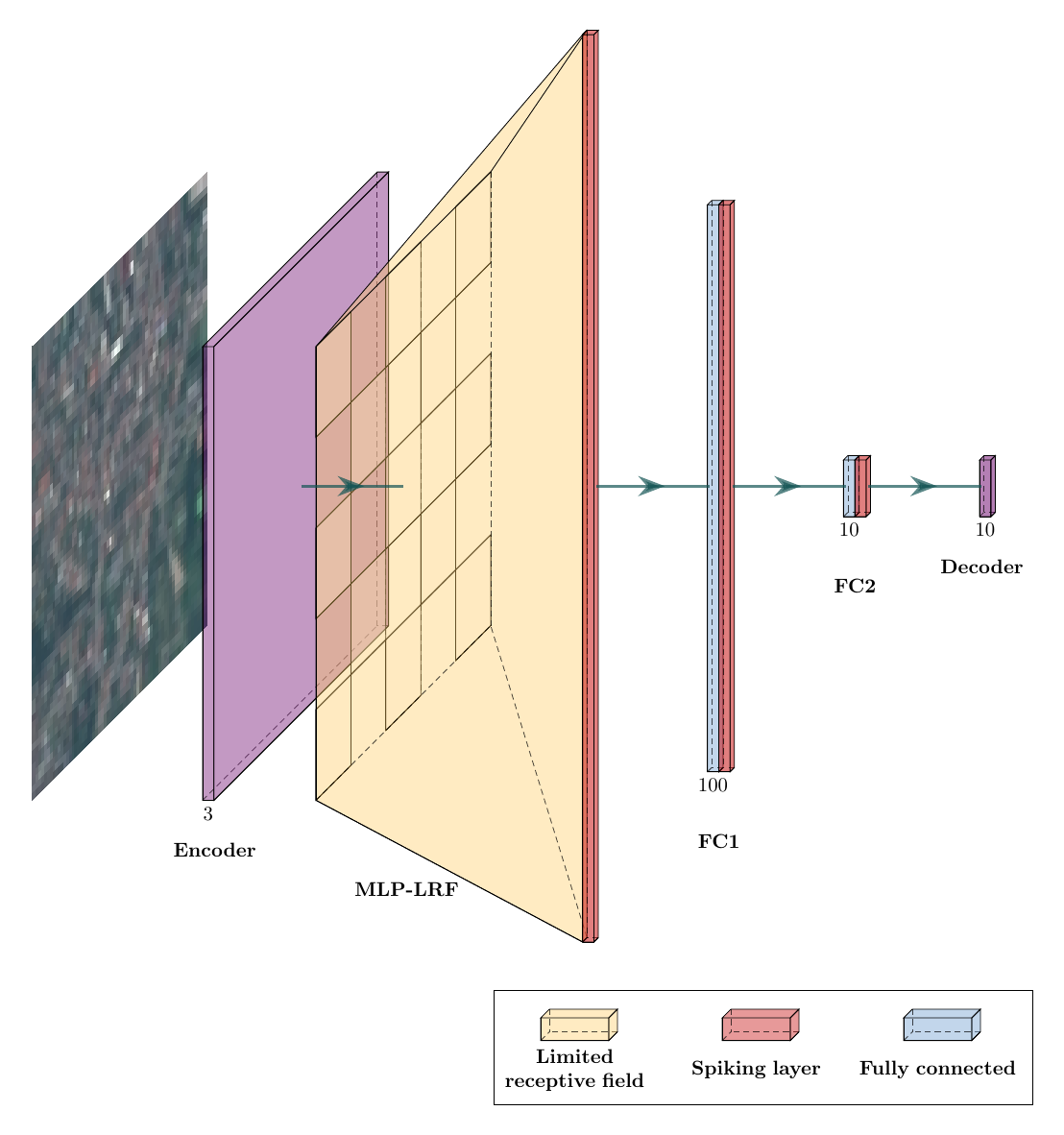}
  \caption{Example of MLP model with LCL. Input is divided in independent limited receptive fields, each one mapped to a MLP layer of spiking neurons.}\label{fig:model_mlplrf}
\end{figure}

Despite the effectiveness of convolution, most of neuromorphic hardware currently available does not support it. 
For instance, mixed-signals chips as BrainScale2 \cite{pehle_brainscales-2_2022} are incompatible with convolution, given the design of the analog crossbar array hardware performing synaptic operations.
In these cases, to infer convolutional models on neuromorphic hardware, it is necessary to unroll the convolutional operation, leading to a memory inefficient implementation.
To emulate the behavior of convolutional models, which process images with a limited receptive fields, we realized Locally Connected Layers in MLP by multiplying all the weights in a layer by a matrix that masks the connections outside of a specific receptive field.
An example of MLP/LCL model is shown in Fig.~\ref{fig:model_mlplrf}.
Furthermore, to potentially increase the performance, some of the models were tested by making the LIF and LI neurons of each layers trainable. 
To this aim, synaptic and membrane constants were randomly initialized at the beginning of training by using a normalized distribution. 
In case of trainable LIF neurons, threshold voltage were also made trainable and initialized by using a user-specified initial value for each layer. 

\section*{References}
\bibliographystyle{astrobib}
\bibliography{bibliography}

\subsection*{Author biography}

\begin{biography}[lunghi.png]{Paolo Lunghi} Since 2018 is Assistant Professor of Aerospace Systems at the Aerospace Science and Technology Department of Politecnico di Milano. His research focuses on the fields of Autonomous GNC systems for spacecraft, Vision-Based Navigation, Hazard Detection and Avoidance systems. He is especially interested in vision-based navigation and other methods to assist and extend spacecraft capabilities by means of Artificial Intelligence, seeking an end-to-end autonomous GNC chain. Email: \texttt{paolo.lunghi@polimi.it}
\end{biography}

\begin{biography}[silvestrini.PNG]{Stefano Silvestrini} is Assistant Professor of Flight Mechanics at the Aerospace Science and Technology Department of Politecnico di Milano. His research interests include the development of Artificial Intelligence algorithms for autonomous GNC in distributed space systems and proximity operations, particularly tailored for embedded applications in small platforms. E-mail: \texttt{stefano.silvestrini@polimi.it}
\end{biography}

\begin{biography}[dominik.png]{Dominik Dold} is a Research Fellow in AI at the European Space Agency's Advanced Concepts Team (ACT). Before joining the ACT, he did a PhD at Heidelberg University and the University of Bern working on neuro-inspired AI, neuromorphic computing, and computational neuroscience. Following his PhD, he was a Researcher in Residence at the Siemens AI Lab in Munich, focusing on bridging graph learning, neuromorphic computing, and cybersecurity applications. 
In his research, he investigates how functionality emerges in simple, energy-efficient, and adaptive systems -- be it biological and artificial neural networks, materials, or robot swarms.  E-mail: \texttt{dodo.science@web.de}
\end{biography}

\begin{biography}[meoni.jpg]{Gabriele Meoni} received his MsC degree in Electronic Engineering and his PhD degree in Information Engineering from the University of Pisa respectively in 2016 and in 2020, where he was supervised by Prof. Luca Fanucci. After completing his doctoral studies, from September 2020 to April 2023 he held the position of Internal Research Fellow at the European Space Agency (ESA) (Advanced Concepts Team (ACT) September 2020 - August 2021, $\Phi$-lab September 2021 - April 2023), where he conducted research on Artificial Intelligence (AI) and neuromorphic computing for onboard spacecraft applications. During 2022-2023, he was a visiting researcher at AI Sweden, focusing on distributed edge learning for satellite constellations. From May 2023 to April 2024, he served as an Assistant Professor in the Faculty of Aerospace Engineering at Delft University of Technology. Currently, Meoni is an Innovation Officer at ESA, with research interests spanning satellite onboard processing, AI for Earth Observation, and neuromorphic computing. Meoni coauthored more than 40 scientific publications. Email: \texttt{gabriele.meoni@esa.int}
\end{biography}

\begin{biography}[hadjiivanov.png]{Alexander Hadjiivanov} is a Research Fellow in AI with the Advanced Concepts Team of the European Space Agency. He received his PhD in AI from the University of New South Wales in Sydney, Australia, with a focus on neuroevolution and bioinspired computation. He is currently working on biorealistic models of the retina, adaptive neuron models and continual learning. His research objective is to incorporate adaptation and homeostasis at all levels of the neural processing pipeline -- from perception to computation -- in order to enable artificial neural networks to continually learn in uncertain environments. Email: \texttt{alexander.hadjiivanov@esa.int}
\end{biography}

\begin{biography}[izzo.jpg]{Dario Izzo} is the Scientific Coordinator of ESA's Advanced Concepts Team. He devised and managed the Global Trajectory Optimization Competitions events, the ESA Summer of Code in Space, and the Kelvins innovation and competition platform. He published more than 200 scientific papers pioneering the use of AI techniques in different field of spacecraft dynamics, guidance and control. He is also the recipient of the Humies Gold Medal for the creation of an AI system having human-competitive capabilities in the design interplanetary trajectories (2013), led the team winning the 8th edition of the Global Trajectory Optimization Competition (2015) and received the Barry Carlton Award (2024). Email: \texttt{dario.izzo@esa.int}
\end{biography}

\end{document}